\documentclass[jmse,review,accept,pdftex,moreauthors]{mdpi}

\firstpage{1} 
\pubvolume{14}
\issuenum{1}
\articlenumber{59}
\pubyear{2026}
\copyrightyear{2025}
\externaleditor{Jianjun Ni }
\datereceived{4 December 2025 }
\daterevised{26 December 2025 }
\dateaccepted{27 December 2025  }
\datepublished{29 December 2025}

\usepackage{makecell}
\renewcommand{\hl}[1]{#1}
\Title{{A} Comprehensive Review of Bio-Inspired Approaches to Coordination, Communication, and System Architecture in Underwater Swarm Robotics}

\Author{%
\hl{Shyalan Ramesh} $^{1,}$*\orcidA{}, 
Scott Mann $^{1}$\orcidB{} and 
Alex Stumpf $^{2}$\orcidC{}}

\AuthorNames{Shyalan Ramesh, Scott Mann and Alex Stumpf}

\address{%
$^{1}$ \quad Department of Computer Science and Information Technology, School of Computing, Engineering, and Mathematical Sciences, La Trobe University, Melbourne, VIC 3086, Australia  
\\
$^{2}$ \quad Department of Engineering, School of Computing, Engineering, and Mathematical Sciences, La Trobe University, Melbourne, VIC 3086, Australia}

\corres{Correspondence: \hl{shyalan.ramesh}@latrobe.edu.au 
}

\abstract{The increasing complexity of marine operations has intensified the need for intelligent robotic systems to support ocean observation, exploration, and resource management. Underwater swarm robotics offers a promising framework that extends the capabilities of individual autonomous platforms through collective coordination. Inspired by natural systems, such as fish schools and insect colonies, bio-inspired swarm approaches enable distributed decision-making, adaptability, and resilience under challenging marine conditions. Yet research in this field remains fragmented, with limited integration across algorithmic, communication, and hardware design perspectives. This review synthesises bio-inspired coordination mechanisms, communication strategies, and system design considerations for underwater swarm robotics. It examines key marine-specific algorithms, including the Artificial Fish Swarm Algorithm, Whale Optimisation Algorithm, Coral Reef Optimisation, and Marine Predators Algorithm, highlighting their applications in formation control, task allocation, and environmental interaction. The review also analyses communication constraints unique to the underwater domain and emerging acoustic, optical, and hybrid solutions that support cooperative operation. Additionally, it examines hardware and system design advances that enhance system efficiency and scalability. A multi-dimensional classification framework evaluates existing approaches across communication dependency, environmental adaptability, energy efficiency, and swarm scalability. Through this integrated analysis, the review unifies bio-inspired coordination algorithms, communication modalities, and system design approaches. It also identifies converging trends, key challenges, and future research directions for real-world deployment of underwater{\linebreak} swarm systems.}


\begin{document}

\section{Introduction}
\textls[-10]{\hl{The} sustainable management of marine resources remains one of the most pressing global challenges, requiring effective monitoring and exploration of ocean \mbox{environments~\cite{di_ciaccio2021monitoring, franke2020operationalizing}}}. Over recent decades, Marine Robotic Systems (MRSs), including Autonomous Underwater Vehicles (AUVs), Remotely Operated Vehicles (ROVs), and 
Unmanned Surface Vessels (USVs) have revolutionised ocean observation and exploration~\cite{petillot2019underwater, zhao2025bioinspired}. A significant trend has been the move from single-vehicle autonomy to collective autonomy, where multiple vehicles cooperate to enhance spatial coverage, robustness, and 
mission~adaptability.

Underwater robotic systems have evolved from early tethered Remotely Operated Vehicles used for inspection and intervention to increasingly autonomous platforms capable of long-duration sensing and decision-making. Today, AUVs, ROVs, and 
USVs are deployed across a wide range of missions, including seabed mapping, habitat and water-quality monitoring, offshore infrastructure inspection, search-and-recovery, and 
operations in remote environments such as under-ice and deep-sea regions. This progression towards higher autonomy and broader deployment has naturally motivated research into multi-robot and swarm systems, where cooperation can improve coverage, redundancy, and 
mission resilience in challenging ocean conditions~\cite{petillot2019underwater, zhao2025bioinspired}.

Swarm robotics draws inspiration from natural collective behaviours observed in fish schools, bird flocks, and 
insect colonies, achieving complex coordination through simple local interactions~\cite{brambilla2013swarm}. Applying these principles to underwater domains introduces distinctive challenges including constrained communication bandwidth, acoustic delay, and 
limited power capacity, but 
it also creates opportunities for innovation~\cite{ismail2021systematic, connor2021current}. Bio-inspired strategies are particularly relevant in marine environments, as 
aquatic organisms have evolved to thrive under precisely the constraints that limit robotic systems: restricted communication, positional uncertainty, and 
energy scarcity~\cite{zhao2025bioinspired}. Fish schools, for 
instance, maintain formation through local perception rather than global positioning~\cite{chen2023modelling}, providing direct insight into efficient coordination for robotic~swarms.

Despite notable progress in individual marine robotic platforms, the 
field of underwater swarm robotics remains conceptually and methodologically fragmented. Extensive research has addressed aspects of swarm coordination, communication, and 
bio-inspired design, yet no unified framework integrates these dimensions into a coherent body of knowledge~\cite{dorigo2021swarm, connor2021current}.

This review addresses this gap through four contributions. First, it unifies underwater swarm robotics across three coupled layers: bio-inspired coordination and optimisation, underwater communication constraints, and 
system architecture, so that algorithms are discussed alongside the physical and networking limitations that shape real deployments. Second, it focuses on marine-grounded bio-inspired methods and their underwater adaptations, using representative exemplars including AFSA, WOA, CRO, and 
MPA to illustrate common design patterns and evaluation considerations. Third, it introduces a four-dimensional classification framework based on communication dependency, environmental adaptability, energy efficiency, and 
swarm scalability, to 
position and compare approaches consistently. Fourth, it synthesises recurring evaluation gaps and open challenges to support more reproducible benchmarking and more informed algorithm selection. In contrast to many existing surveys that treat bio-inspired algorithms broadly across general robotics and to
surveys that concentrate on narrower subsets of underwater coordination techniques, this survey is explicitly underwater-specific and integrates algorithms, communications, and 
architecture into a single \mbox{comparative~framework.}

\begin{figure}[H]\hspace{-1.3mm}
\includegraphics[width=1\linewidth]{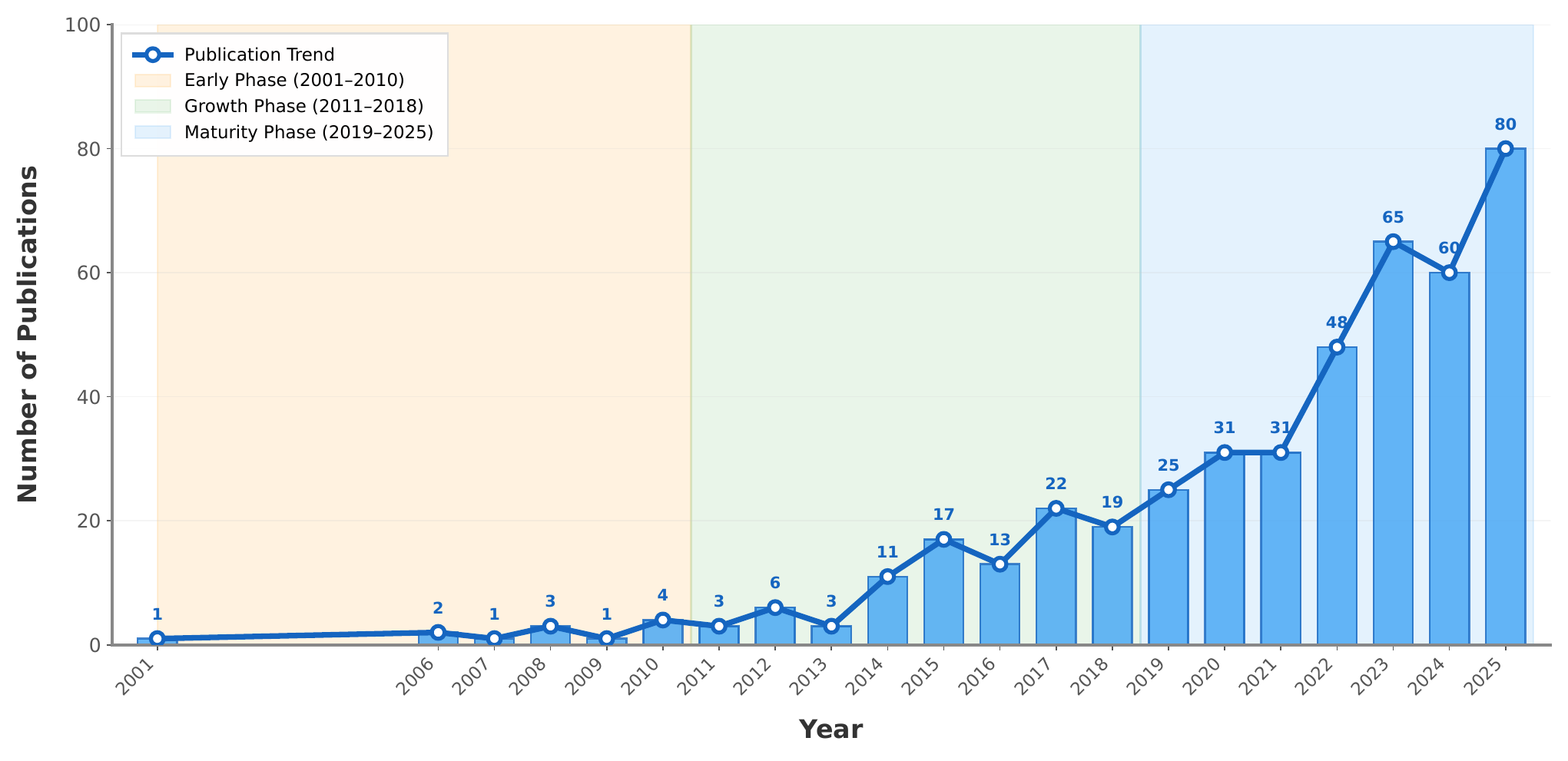}
\caption{\hl{Publication} 
 trend for underwater swarm robotics research (2001 to 2025). The analysis includes 446 research articles from major databases, demonstrating exponential growth in the field. Search keywords included: marine swarm robotics, AUV swarm robotics, underwater swarm robotics, and 
related~terms.}
\label{fig:publication_trend}
\end{figure}

To contextualise the scope and evolution of this field, Figure~\ref{fig:publication_trend} presents the publication trend for underwater swarm robotics research from 2001 to~2025.

\subsection*{\hl{Review~Methodology} 
}

\begin{figure}[H]\hspace{-1.3mm}
\includegraphics[width=1\linewidth]{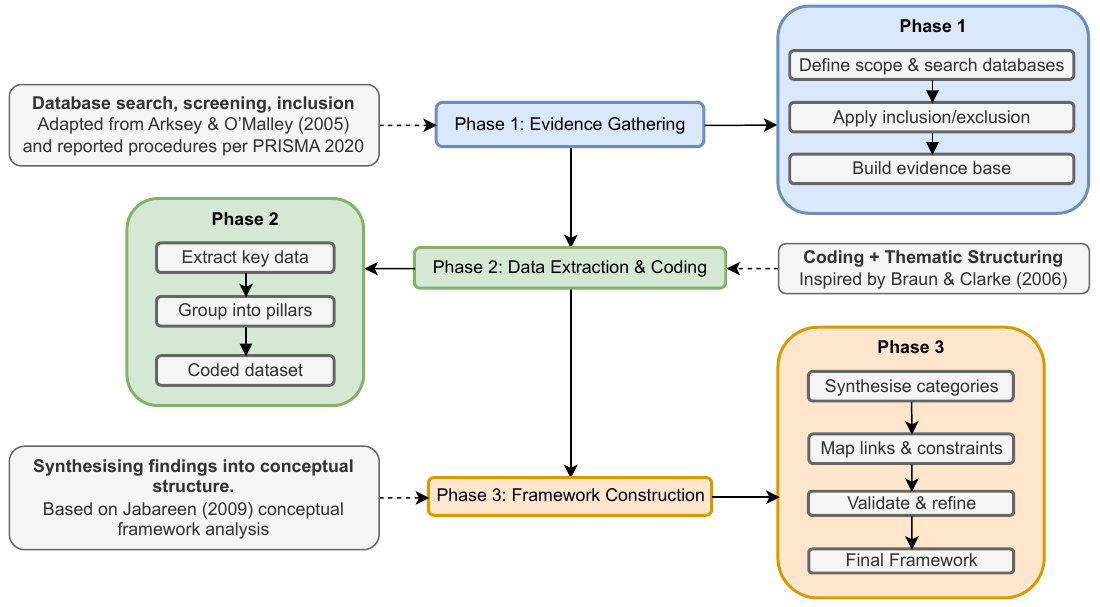}
\caption{\hl{Three-phase} 
research methodology for synthesising the literature on bio-inspired underwater swarm robotics. The methodology integrates evidence gathering~\cite{arksey2005scoping}, systematic data extraction and coding~\cite{braun2006thematic}, and 
integrated synthesis~\cite{jabareen2009building} to consolidate findings across algorithmic, communication, and 
system design~domains.}
\label{fig:methodology}
\end{figure}

This study adopts a structured narrative review design guided by a transparent and replicable search and screening process. Figure~\ref{fig:methodology} illustrates the three-phase methodological approach, which combines scoping review principles, thematic analysis, and 
conceptual framework~development.

Bibliographic records were collected from seven major scholarly databases (MDPI, OpenAlex, Scopus, Springer, ScienceDirect, IEEE, and 
Google Scholar) using combinations of keywords related to underwater swarm robotics and marine multi-robot systems. These databases were chosen because they are widely used indexing sources spanning robotics, marine engineering, and 
communications; however, we do not claim a precise percentage coverage of all relevant publications due to overlap and variability in indexing across venues. Keyword searches employed Boolean logic combining terms such as (``underwater'' OR ``marine'') AND (``swarm'' OR ``multi-robot'' OR ``multi-AUV'') AND (``robotics'' OR ``coordination'' OR ``cooperation''), along with specific phrases including ``marine swarm robotics'', ``AUV swarm robotics'', ``underwater swarm robotics'', and 
related terms. All searches were restricted to English-language publications. The search window spanned publications from 2001 to 2025. Only peer-reviewed research articles were included; review papers, conference proceedings, theses, and 
non-peer-reviewed sources were~excluded. 

Records were screened to remove duplicates using DOI matching and title comparison, and 
to retain only studies that addressed underwater or marine swarm robotics, multi-AUV cooperation, or 
closely related multi-vehicle coordination problems in aquatic environments. Approximately 65\% of the papers from the original sources were excluded during the filtering and consolidation process, which removed review papers, conference papers, and 
duplicate~publications.

The final corpus comprises 446 peer-reviewed research articles. Titles, abstracts, and, where necessary, full texts were examined to confirm relevance and to assign each article to one or more thematic categories reflecting its primary contribution: bio-inspired coordination and control algorithms, underwater communication and networking strategies, and 
system design and implementation (including platforms, sensing, energy management, and 
validation). This thematic coding underpins the organisation of the review: Section~\ref{foundations} introduces the foundational context, Sections~\ref{coordination}--\ref{systemdesign} synthesise coordination, communication, and 
system design themes, and 
Section~\ref{framework} presents an integrative classification that links algorithmic, communication, and 
hardware perspectives across the corpus. It is important to acknowledge limitations: the dataset may have gaps as some journals and sources may have been missed, and 
the search strategy may not capture all relevant publications using alternative~terminology.

\section{Foundations of Marine Swarm~Robotics}
\label{foundations}

Underwater swarm robotics operates within a uniquely challenging domain that differs fundamentally from terrestrial and aerial environments. Understanding these challenges, the 
governing principles of swarm intelligence, and 
the relevance of bio-inspired design is essential for developing effective and sustainable Marine Robotic Systems. This foundational context establishes the environmental and conceptual constraints within which bio-inspired coordination, communication, and 
system design approaches \mbox{must~operate}.

\subsection{Characteristics and Challenges of the Underwater~Environment}
The underwater environment imposes physical, acoustic, and 
operational constraints that significantly influence the design and behaviour of swarm robotic systems. Unlike terrestrial or aerial systems, underwater platforms must operate in a medium that limits communication, localisation, sensing, and 
energy~management.

A primary challenge is navigation. Global Positioning System (GPS) signals do not propagate underwater, forcing AUVs to rely on alternative methods such as Inertial Navigation Systems (INSs), dead reckoning, and 
acoustic positioning arrays~\cite{connor2021current, sorensen2025localization}. State-estimation filters, particularly (extended) Kalman filtering and invariant EKF variants, are widely used to fuse noisy measurements and reduce drift in underwater navigation~\cite{potokar2021invariant, ligeza2023reconstructing}. INS accumulates drift over time, while dead reckoning suffers from bias and integration errors. Acoustic positioning systems, including Long Baseline (LBL) and Ultra-Short Baseline (USBL) methods, provide absolute fixes but are constrained by propagation delays, multipath interference, and 
the need for pre-deployed infrastructure. These uncertainties complicate formation control and coordinated behaviours, as 
individual agents may hold inconsistent estimates of position relative to their~neighbours.

Communication represents another fundamental limitation. Underwater environments preclude the use of conventional Radio Frequency channels. Instead, communication relies primarily on acoustic, optical, or 
low-frequency electromagnetic modalities, each with trade-offs in range, bandwidth, and 
reliability. For example, Awan et~al. reviewed system-level constraints and networking challenges in underwater wireless systems~\cite{awan2019underwater}, Saeed et~al. surveyed underwater optical wireless communications and localisation~\cite{saeed2019underwater}, and 
Alahmad et~al. experimentally evaluated underwater RF/EM transmission for AUV applications~\cite{alahmad2023experimental}. Acoustic links offer long range but suffer from low data rates, latency, and 
ambient noise; optical communication enables high throughput over short distances but is sensitive to turbidity; and electromagnetic transmission is restricted to a few metres due to rapid attenuation in seawater. These limitations necessitate swarm control architectures that tolerate sparse, delayed, and 
intermittent~connectivity.

Environmental variability further compounds these challenges. Variations in temperature, salinity, and 
current affect vehicle dynamics, sensing performance, and 
communication range. High pressures at depth require robust pressure-rated housings, while saltwater corrosion demands protective materials and coatings. Energy management remains a persistent constraint, as 
finite battery capacity limits endurance and dictates careful trade-offs among propulsion, sensing, computation, and 
communication. 

Collectively, the 
absence of global positioning, constrained communication, environmental uncertainty, and 
limited energy define a complex design space for marine swarm robotics. These factors motivate decentralised, adaptive, and 
energy-aware coordination approaches capable of operating reliably with partial and delayed~information.

\subsection{Principles of Swarm~Intelligence}
Swarm intelligence examines how complex collective behaviours emerge from local interactions among many simple agents~\cite{ismail2021systematic}. Natural examples include fish schools, bird flocks, and 
insect colonies, where individuals follow local behavioural rules based on neighbour perception and environmental cues~\cite{brambilla2013swarm}. Without any central controller, such collectives achieve tasks including exploration, foraging, migration, and 
defence.

Key principles of swarm intelligence include decentralisation, scalability, robustness, and 
emergence. In decentralised systems, each agent acts autonomously based on local information, reducing dependency on global knowledge and central coordination. Scalability arises when identical rules can govern both small and large groups without structural modification. Robustness stems from redundancy, where loss or malfunction of individual agents has minimal effect on collective performance. Emergence describes the spontaneous formation of complex, adaptive behaviours from simple rules of~interaction.

In robotic applications, these principles translate into algorithms that govern local perception, control, and 
decision-making. Agents modify their movements and roles according to the states of nearby neighbours, producing emergent phenomena such as formation control, aggregation, and 
exploration. Examples include robotic fish that exploit hydrodynamic interactions to conserve energy~\cite{zhang2024energy}, decentralised binary decision-making strategies that enable collective motion through local perception~\cite{wu2025minimalistic}, and 
pheromone-inspired coordination mechanisms for distributed exploration~\cite{geng2019pheromone, li2019pheromone}. 

These properties are particularly suited to underwater environments. Communication limitations naturally encourage local interactions, while uncertain conditions require adaptive, fault-tolerant behaviours. Swarm intelligence thus provides the conceptual foundation for designing cooperative Marine Robotic Systems that can function effectively with minimal communication and incomplete environmental~awareness.

\subsection{Relevance of Bio-Inspired Design in Marine~Systems}
Bio-inspired design offers a powerful framework for developing underwater swarm robotics, as 
marine organisms have evolved to operate efficiently under the same constraints that challenge robotic systems: restricted communication, positional uncertainty, and 
limited energy availability~\cite{zhao2025bioinspired}. By emulating natural coordination and sensing strategies, robotic systems can achieve resilient, low-energy, and 
adaptive behaviours suited to underwater~conditions.

Marine organisms exemplify efficient coordination mechanisms that require minimal communication. Fish schools maintain cohesive formations through local alignment and spacing rather than global control~\cite{chen2023modelling}, while hydrodynamic coupling between individuals can reduce overall energy expenditure by more than 50\% compared with solitary swimming~\cite{zhang2024energy}. These naturally evolved mechanisms demonstrate how energy-efficient cooperation and adaptive formation control can emerge without continuous data exchange and absolute~positioning.

\begin{figure}[H]\hspace{-1.5mm}
\includegraphics[width=1\linewidth]{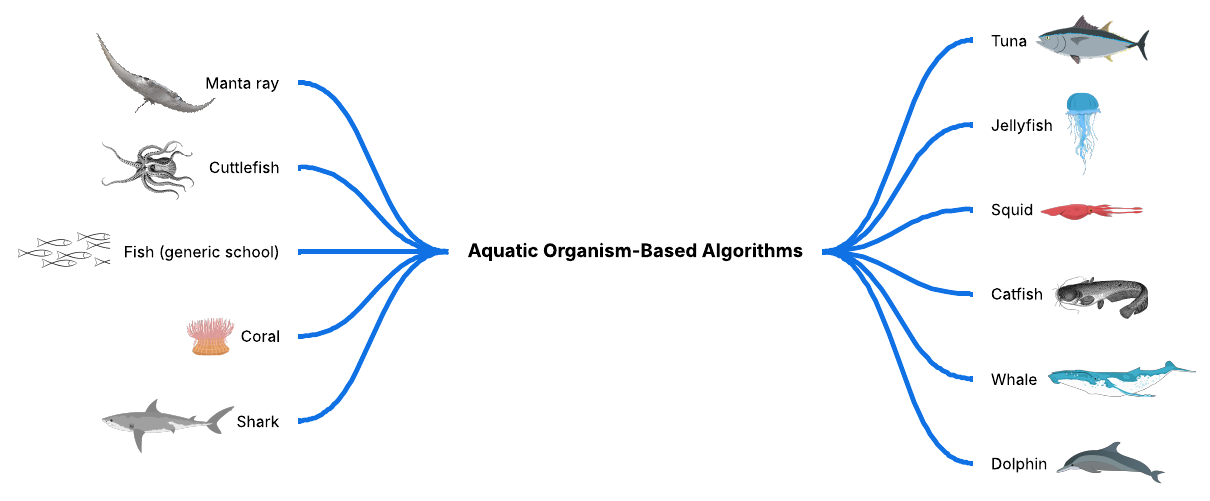}
\caption{Aquatic organisms that underpin many bio-inspired optimisation algorithms, illustrating representative marine species used as the biological basis for these~methods.}
\label{fig:organisms}
\end{figure}

Several marine-specific bio-inspired algorithms capture such biological strategies in computational form. The Artificial Fish Swarm Algorithm (AFSA) models visual perception, swarming, and 
following behaviours~\cite{cai2023cooperative}; the Whale Optimisation Algorithm (WOA) draws from bubble-net hunting patterns of humpback whales~\cite{yan2022woa}; the Coral Reef Optimisation (CRO) algorithm simulates coral reproduction and depredation dynamics~\cite{emami2021cro}; and the Marine Predators Algorithm (MPA) represents predator--prey interactions across distinct velocity phases~\cite{zhao2025bioinspired}. These algorithms are inherently compatible with underwater constraints, requiring limited communication, operating effectively under positional uncertainty, and 
promoting energy-efficient exploration and cooperation. Recent developments have further addressed specific underwater challenges, such as hydrodynamic constraints and irreversible settlement behaviours in strong current environments~\cite{ramesh2025naupliusoptimisationautonomoushydrodynamics}. A wide variety of nature-inspired optimisation algorithms has been proposed~\cite{darvishpoor2023nature}. Figure~\ref{fig:organisms} shows several aquatic organism-based algorithms inspired by marine~species.

\textls[-6]{Beyond optimisation-based bio-inspired algorithms, neural-inspired control paradigms offer complementary approaches. Spinal neural system models have been applied to heterogeneous AUV cooperative hunting~\cite{ni2018improved}, while bio-inspired neural networks have enabled decision-making and motor control in differential robots~\cite{guerrerocriollo2023bioinspired}. Reinforcement learning approaches, such as neural network-based Q-learning for path planning in unknown environments~\cite{ni2016bioinspired}, further extend the scope of bio-inspired methods to learning-based~coordination.}

\newpage
Bio-inspired strategies have been applied to improve communication topologies, formation control, task allocation, and 
environmental interaction within underwater swarm robotics. They promote robustness, scalability, and 
adaptability, qualities essential for long-duration missions in unpredictable marine environments~\cite{kaur2024energy}. Despite growing adoption, systematic guidance for integrating these algorithms with communication protocols and hardware architectures remains limited. This review addresses that gap by synthesising bio-inspired coordination, communication, and 
design approaches into a coherent framework for future intelligent marine robotic~systems.

\section{Bio-Inspired Coordination~Mechanisms}
\label{coordination}

Bio-inspired coordination mechanisms form the core of underwater swarm robotics, enabling multiple autonomous agents to achieve collective objectives through local interactions and distributed decision-making. This section examines the principal coordination strategies derived from marine biological systems. It focuses on how local interactions enable decentralised control, how formation control supports collective motion, how task allocation distributes responsibilities, and 
how collective decision-making emerges from individual behaviours. Throughout, the 
emphasis remains on how biologically inspired rules translate into concrete coordination algorithms suitable for communication-limited, energy-constrained underwater~swarms.

\subsection{Local Interaction and Decentralised~Control}

Decentralised control represents a fundamental principle of swarm robotics, where coordination emerges from local interactions rather than centralised command. This approach is particularly suited to underwater environments, where communication constraints and intermittent connectivity make global coordination~impractical.

Local interaction mechanisms enable agents to coordinate through simple behavioural rules based on neighbour perception and environmental sensing. In biological swarms, individual organisms respond to nearby neighbours' positions and velocities, producing emergent collective patterns without any central controller~\cite{brambilla2013swarm}. Robotic swarms replicate these mechanisms by implementing local control policies that govern perception, movement, and 
decision-making, allowing distributed coordination that scales efficiently with swarm~size.

The Artificial Fish Swarm Algorithm (AFSA) exemplifies local interaction through its visual perception model, where artificial fish respond to neighbours within a defined visual range. Adeli provided an early review of AFSA methods and applications~\cite{adeli2012review}, Pourpanah et~al. summarised more recent AFSA variants and application directions~\cite{pourpanah2022afsa}, and 
Peraza et~al. presented an AFSA-based approach in a computational-intelligence setting~\cite{peraza2022afsa}. Each agent adjusts its position based on local stimuli such as prey attraction, swarming, and 
following behaviours. This mechanism enables coordination through environmental sensing rather than explicit message exchange, making AFSA highly compatible with low-bandwidth acoustic communication~environments. 

\begin{figure}[H]\hspace{-1.5mm}
\includegraphics[width=0.75\linewidth]{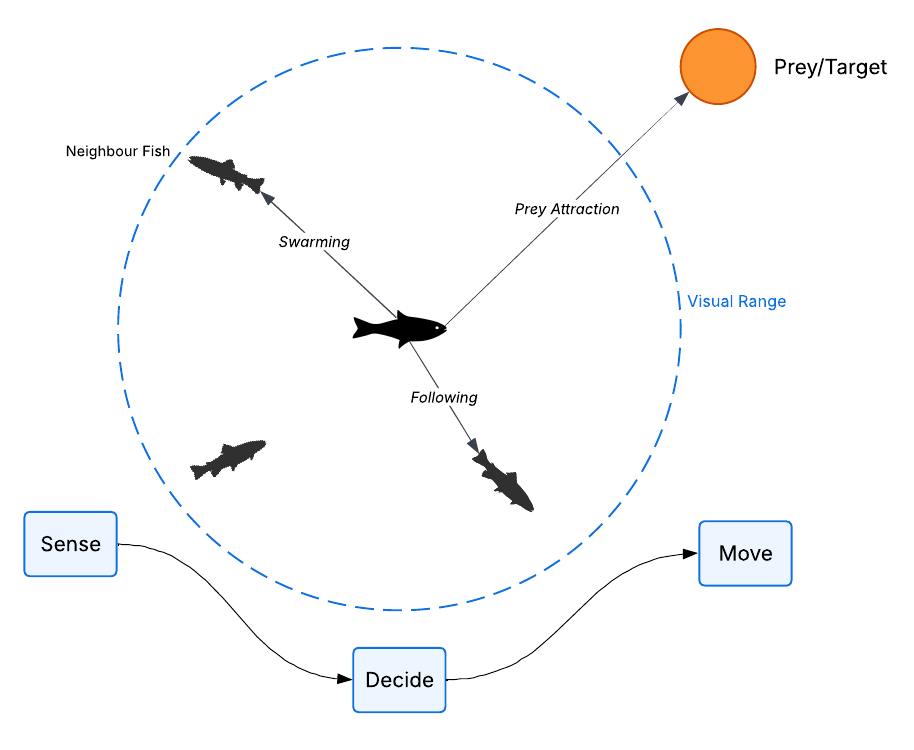}
\caption{Visual perception and behaviour selection in the Artificial Fish Swarm Algorithm (AFSA). The central artificial fish perceives neighbouring agents and environmental stimuli within its visual range and dynamically selects between prey attraction, swarming, and 
following behaviours based on local sensory input. Coordination emerges from environmental perception rather than direct~communication.}

\label{fig:afsa}

\end{figure}

Figure~\ref{fig:afsa} presents a conceptual illustration of the AFSA visual perception and behaviour selection mechanism, showing how artificial fish perceive nearby individuals and environmental stimuli to dynamically select between prey attraction, swarming, and 
following~behaviours.

Decentralised control architectures eliminate single points of failure and enhance resilience. Adaptive coordination strategies allow vehicles to adjust parameters in response to varying communication quality, enabling continuity despite packet loss and latency. Minimalistic binary coordination strategies also exemplify effective decentralisation: agents make binary rotation decisions (for example, left or right) based solely on neighbour headings~\cite{wu2025minimalistic}. Such strategies achieve coherent collective motion through local perception, requiring no explicit synchronisation or global~positioning.

Local interaction mechanisms thus provide robustness, scalability, and 
adaptability. As swarm size increases, the 
same local rules govern behaviour without structural modification, allowing the swarm to maintain coordination performance across a wide range of operational~scales.

\subsection{Formation Control and Collective~Motion}
Formation control enables underwater swarms to maintain desired spatial configurations while navigating dynamic marine environments. Bio-inspired formation control strategies draw directly from natural collective behaviours, particularly those observed in fish schools, where individuals sustain relative positions through local alignment and spacing~\cite{chen2023modelling}.

Fish schooling provides a natural template for energy-efficient collective motion. Experiments demonstrate that fish reduce energy expenditure per tail beat by more than 50\% compared with solitary individuals, benefiting from hydrodynamic interactions~\cite{zhang2024energy}. These mechanisms illustrate how coordinated motion can emerge without continuous data exchange and precise global positioning, providing models for underwater swarm formation~control.

\begin{figure}[H]\hspace{-1.5mm}
\includegraphics[width=0.8\linewidth]{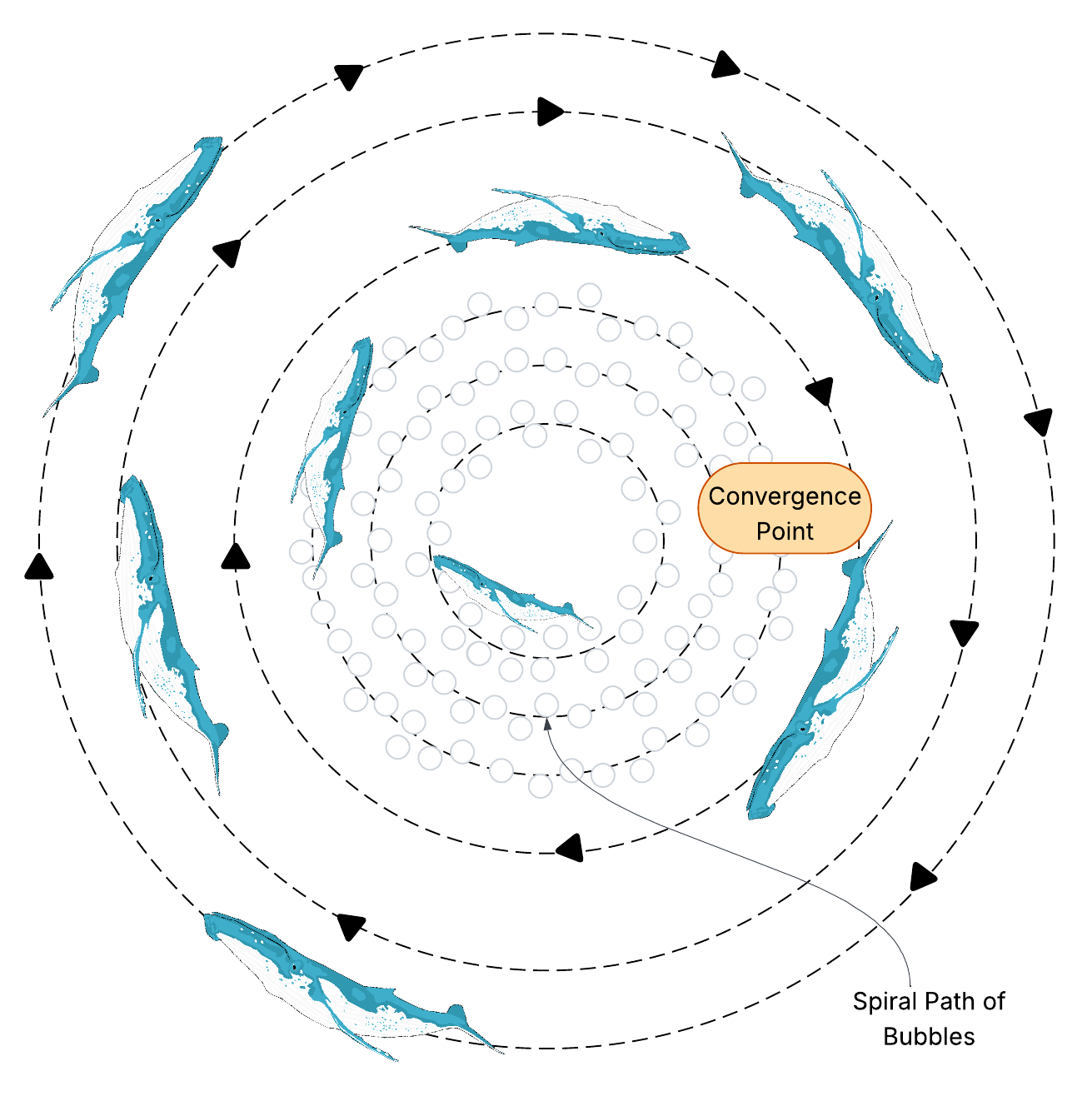}
\caption{Bubble-net feeding behaviour of humpback whales illustrating the search mechanism of the Whale Optimisation Algorithm (WOA).}
\label{fig:woa}
\end{figure}

WOA models the cooperative bubble-net feeding behaviour of humpback whales, in 
which individuals encircle and herd prey through coordinated movement patterns~\cite{mirjalili2016woa, yan2022woa}. This strategy underpins the algorithm’s encircling mechanism, representing agents that maintain adaptive relative positions within the search space, while the spiral bubble-net model guides convergence towards promising regions, analogous to coordinated manoeuvring observed in nature. Rana et~al.~\cite{rana2020whale} further analysed these dynamics, highlighting the balance between exploration and exploitation achieved through such collective strategies. The algorithmic abstraction of this behaviour is illustrated in Figure~\ref{fig:woa}.

\begin{figure}[H]
\begin{adjustwidth}{-\extralength}{0cm}
\centering
\includegraphics[width=\linewidth]{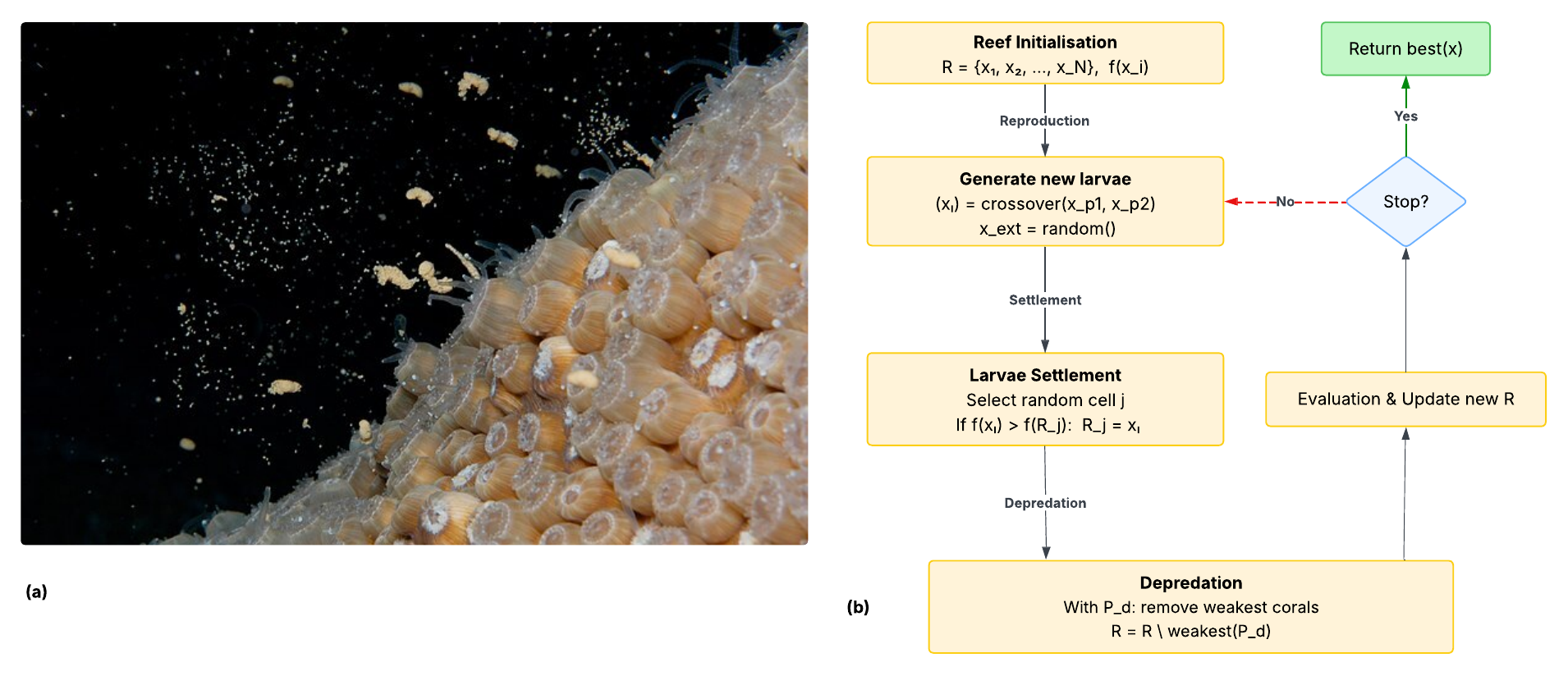}
\end{adjustwidth}
\caption{\hl{Coral} 
 Reef Optimisation (CRO) algorithm: (\textbf{a}) coral spawning illustrating the biological inspiration; (\textbf{b}) flowchart of the CRO process showing reef initialisation, larvae generation and settlement, evaluation, and 
depredation~steps.}
\label{fig:cro}
\end{figure}

CRO provides an alternative bio-inspired optimisation framework grounded in the competitive dynamics of coral reefs, where individual corals occupy spatial positions and interact through spawning and larvae settlement~\cite{salcedo2014cro}. The algorithm models the ecological processes of reproduction, settlement, and 
depredation on a discretised reef grid, balancing exploration and exploitation through spatial competition. This distributed mechanism promotes redundancy and resilience against individual failure, enabling adaptive population evolution over time. The biological inspiration and algorithmic process of CRO are illustrated in Figure~\ref{fig:cro}. Figure~\ref{fig:cro}b summarises the CRO iteration on a discretised reef, where $R=\{R_1,\dots,R_N\}$ denotes reef cells storing candidate solutions and $f(\cdot)$ is the objective (fitness) to be maximised. New larvae $x'$ are generated from parent solutions (e.g., via crossover and random injection $x_{\text{ext}}$ for exploration), then a larva attempts to settle into a randomly chosen cell $j$. If it improves the resident solution, i.e.,~$f(x')>f(R_j)$, it replaces it, implementing local competition for space. Periodically, depredation removes a fraction of the weakest corals with probability $P_d$, preventing stagnation and promoting exploration; the loop repeats until a stopping criterion is met and the best coral is returned. Minimalistic formation-control approaches based on binary interactions and local decision rules likewise demonstrate cohesive group motion under limited communication: each AUV responds to neighbour headings with simple binary decisions (e.g., rotate left or right by a fixed angle), enabling emergent coherent collective motion without predefined global targets~\cite{wu2025minimalistic}.

Underwater formation control must accommodate GPS denial and positional uncertainty. Algorithms based on relative positioning naturally address these challenges, mirroring biological systems that rely on neighbour-relative cues rather than absolute global references. Consequently, relative localisation and neighbour-awareness form the foundation of practical marine swarm formation~strategies.

\subsection{Task Allocation and Role~Differentiation}
Task allocation distributes mission responsibilities among swarm members to maximise efficiency and exploit heterogeneity in agent capability. Bio-inspired task allocation draws from natural systems in which individuals assume roles according to local context and resource~availability.

The Coral Reef Optimisation algorithm embodies adaptive task allocation through competitive reef dynamics, where corals vie for space based on fitness values~\cite{salcedo2014cro, emami2021cro}. Its depredation mechanism removes weaker solutions, creating natural role differentiation as stronger agents assume more critical responsibilities while others are replaced and reassigned. This dynamic reassignment provides inherent fault tolerance and adaptability to changing~conditions.

\begin{figure}[H]\hspace{-1.5mm}
\includegraphics[width=0.85\linewidth]{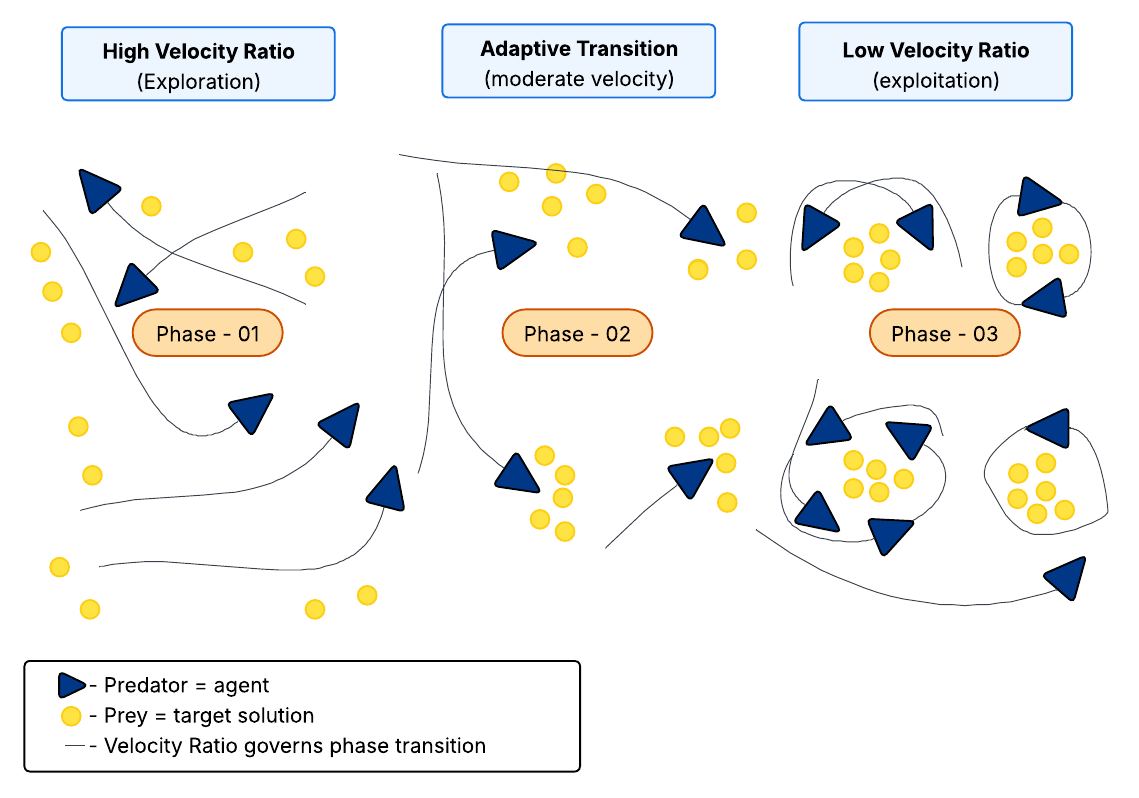}
\caption{Conceptual illustration of the Marine Predators Algorithm (MPA) showing the three-phase optimisation process governed by predator--prey velocity ratios. The schematic depicts the transition from high-velocity exploration to low-velocity exploitation, capturing the algorithm’s adaptive search~dynamics.}
\label{fig:mpa}
\end{figure}

The Marine Predators Algorithm (MPA) models cooperative predator--prey interactions governed by relative velocity ratios~\cite{faramarzi2020mpa, albetar2023mpa}. Agents alternate between exploration and exploitation phases, acting as explorers at high-velocity ratios and as refiners at low-velocity ratios. This phase-based transition enables adaptive task allocation that responds dynamically to mission progress and environmental complexity. Improved modelling of velocity ratio adaptation and foraging dynamics has further enhanced the algorithm’s search balance~\cite{chun2024improved}. The overall three-phase optimisation process is illustrated conceptually in Figure~\ref{fig:mpa}.

AFSA supports task allocation through local perception, where each agent selects targets based on local environmental information rather than central assignment~\cite{cai2023cooperative}. This method aligns well with underwater conditions, allowing distributed agents to self-select tasks based on sensory input and spatial~proximity.

Role differentiation can also be extended to heterogeneous swarms, where agents with different sensing and mobility capabilities assume complementary functions. Behaviour-driven frameworks enable dynamic alliance formation based on task needs and agent attributes~\cite{liang2020behavior}. Such frameworks integrate global mission objectives with local autonomy, ensuring coordination even under uncertainty and communication~delays.

\subsection{Collective Decision-Making and~Learning}
Collective decision-making allows underwater swarms to achieve consensus and coordinated action through local interaction and distributed information exchange. Bio-inspired collective decisions emerge without explicit global control, relying instead on gradual convergence of local~preferences.

The Marine Predators Algorithm demonstrates distributed decision-making through its three-phase optimisation process, where agents collectively transition between exploration and exploitation~\cite{faramarzi2020mpa}. The algorithm’s dynamics enable agents to identify promising areas cooperatively and converge towards shared targets. Similarly, the 
Whale Optimisation Algorithm achieves collective consensus as agents synchronise their movement towards the globally best position through its encircling behaviour~\cite{mirjalili2016woa, rana2020whale}.

Learning mechanisms further enhance collective coordination by enabling adaptation based on environmental feedback. While most marine-specific bio-inspired algorithms employ fixed behavioural rules, hybrid approaches that integrate reinforcement learning and adaptive parameter tuning could yield improved responsiveness and autonomy~\cite{zhao2025bioinspired}. Neural network-based Q-learning has been demonstrated for robot path planning in unknown environments~\cite{ni2016bioinspired}, and 
spinal neural system approaches have enabled coordinated hunting behaviours among heterogeneous AUVs~\cite{ni2018improved}. Recent work on neuromodulation for motor control further illustrates the potential of bio-inspired neural architectures for robotic decision-making~\cite{guerrerocriollo2023bioinspired}. These adaptive extensions would allow swarms to evolve coordination strategies in response to environmental variability, communication degradation, and 
mission~progress.

Underwater collective decision-making must function under delayed, limited communication. Algorithms that propagate decisions through local neighbour interactions, rather than global synchronisation, offer a robust solution. Such distributed learning and consensus mechanisms represent a key direction for improving the adaptability of future marine robotic~swarms.

\subsection{Hybrid Coordination~Approaches}

Recent research has explored hybrid coordination algorithms that integrate multiple algorithmic principles to overcome limitations of single-paradigm systems. Table~\ref{tab:hybrid_underwater_algorithms} summarises representative hybrid methods applied in underwater swarm robotics, combining swarm intelligence, graph-based reasoning, and 
control-theoretic approaches. A comprehensive comparative evaluation of marine-specific algorithms is presented in Section~\ref{framework}.

\begin{table}[H]
\caption{Recent hybrid algorithms applied in underwater swarm~robotics.}
\label{tab:hybrid_underwater_algorithms}
\small

\begin{adjustwidth}{-\extralength}{0cm}
\begin{tabularx}{\linewidth}{m{5.9cm}<{\raggedright} c c l c c}
\toprule
\textbf{Algorithm} & \textbf{Environment} & \textbf{{\#} 
 Robots} & \textbf{Application Domain} & \textbf{Verification} & \textbf{Year} \\
\midrule
IMOPSO--DWA hybrid algorithm~\cite{sun2025multi} & 3D & Multiple & Cooperative path planning & Simulation & 2025 \\
\midrule
Neural leader--follower control~\cite{praczyk2025neural} & \makecell{2D\\(with depth)} & \makecell{5 (1 leader\\+ 4 followers)} & Formation control & Simulation & 2025 \\
\midrule
Hybrid formation control (leader--follower)~\cite{zhang2023hybrid} & 3D & \makecell{7 (1 leader\\+ 6 followers)} & Formation control & Simulation & 2023 \\
\midrule
Fast Graph Pointer Network (FGPN)~\cite{ru2023fgpn} & 2D & Multiple & Task assignment & Simulation & 2023 \\
\midrule
PSO + ELKAI solver~\cite{mu2025coverage} & 3D & Multiple & Coverage path planning & Simulation & 2025 \\
\midrule
HAP--IAPF (Hybrid APF)~\cite{zhao2022cooperative} & 3D & 8 & Cooperative hunting & Simulation & 2022 \\
\bottomrule
\end{tabularx}
\end{adjustwidth}

\noindent{\footnotesize{%
\hl{Sources}: 
 Sun et~al. (2025)~\cite{sun2025multi}, Praczyk (2025)~\cite{praczyk2025neural}, Zhang et~al. (2023)~\cite{zhang2023hybrid}, Ru et~al. (2023)~\cite{ru2023fgpn}, Mu et~al. (2025)~\cite{mu2025coverage}, Zhao et~al. (2022)~\cite{zhao2022cooperative}.
}}
\end{table}

\section{Underwater Communication~Strategies}
\label{communication}

Underwater communication systems form the backbone of swarm coordination, enabling information exchange among autonomous agents operating in complex and variable marine environments. Unlike terrestrial or aerial systems, underwater communication must contend with physical constraints that severely limit bandwidth, range, and 
reliability. This section reviews the principal communication modalities used in Marine Robotic Systems, examines network topologies that support swarm connectivity, explores bio-inspired communication protocols, and 
discusses mechanisms for delay tolerance, fault handling, and 
synchronisation. These communication layers provide the physical substrate and performance envelope within which bio-inspired coordination and signalling schemes must be implemented in~practice.

\subsection{Acoustic, Optical, and Hybrid Communication~Modalities}
Underwater communication primarily employs three physical modalities (acoustic, optical, and 
electromagnetic), each with distinct performance characteristics governed by signal propagation in seawater. Acoustic communication remains the dominant technology due to its long range and broad applicability for Marine Robotic Systems~\cite{gussen2021optimisation}. Acoustic waves propagate at approximately 1500~m/s, enabling links from tens of metres to several kilometres, but 
with limited bandwidth and strong susceptibility to multipath, Doppler, and 
environmental noise~\cite{wubben2020challenges}. Table~\ref{tab:communication_tech} summarises the comparative characteristics of these technologies; for electromagnetic links in water, propagation speed is approximated as $v \approx c/n$ with representative seawater $n \approx 1.33$~\cite{quan1995refractive}. Note that both optical and RF are electromagnetic communications operating at different frequency ranges; optical uses blue--green wavelengths (450--550~nm), while RF uses lower frequencies (kHz--GHz).

\begin{table}[H]
\caption{Comparative characteristics of underwater communication~technologies.}
\label{tab:communication_tech}
\footnotesize

\begin{adjustwidth}{-\extralength}{0cm}
\setlength{\tabcolsep}{1.8mm}
\begin{tabularx}{\linewidth}{%
m{3.3cm}<{\raggedright}%
m{2.3cm}<{\raggedright}%
m{1.4cm}<{\raggedright}%
m{2.1cm}<{\raggedright}%
m{1.6cm}<{\centering}%
m{2.5cm}<{\raggedright}%
m{2.8cm}<{\raggedright}}
\toprule
\textbf{Communication Type} & \textbf{Typical Range} & \textbf{Data Rate} & \textbf{Propagation Speed} & \textbf{Energy Efficiency} & \textbf{Environmental Sensitivity} & \textbf{Typical Use Case} \\
\midrule
Acoustic & 100~m to 20~km & 10~bps to 100~kbps & 1500~m/s & Moderate & High (temperature, salinity) & Long-range swarm coordination \\
\midrule
Optical (EM) & 1 to 100~m & 1~Mbps to 1~Gbps & $\approx$$2.25 \times 10^8$~m/s & High \mbox{(short-range)} & Very high (turbidity, scattering) & Data exchange between nearby~AUVs \\
\midrule
Electromagnetic (RF) & <2~m & 100~kbps to 10~Mbps & $\approx$$2.25 \times 10^8$~m/s & Low & High (attenuation) & Near-surface communication \\
\midrule
Magnetic Induction (MI) & Short-to-medium & High bandwidth & Negligible delay & High & Low (predictable channel) & Short-range control signalling \\
\midrule
\textls[-15]{Hybrid (Acoustic + Optical)} & \textls[-25]{Context-dependent} & Adaptive & Adaptive & Adaptive & Adaptive & Multi-modal swarm communication \\
\bottomrule
\end{tabularx}
\end{adjustwidth}

\noindent{\footnotesize{%
\hl{Sources}: 
Adapted from Li et~al. (2025)~\cite{li2025recent}, Wubben et~al. (2020)~\cite{wubben2020challenges}, Gussen et~al. (2021)~\cite{gussen2021optimisation}, Vali et~al. (2025)~\cite{vali2025survey}, Wang et~al. (2022)~\cite{wang202215}, Busacca et~al. (2024)~\cite{busacca2024adaptive}, and 
Quan and Fry (1995)~\cite{quan1995refractive}.
}}
\end{table}

Recent developments in underwater wireless communication have expanded AUV swarm capabilities, but 
the aquatic environment still imposes severe constraints. Acoustic communication is the dominant long-range modality for AUV swarms~\cite{gussen2021optimisation}, enabling essential functions like localisation and information sharing. However, acoustic channels are limited to roughly 10~kbaud data rates~\cite{wubben2020challenges} and suffer from high latency and noise, making coordination challenging in large swarms~\cite{pal2022communication}.

As a complement to acoustics, optical and RF methods are being explored. Optical systems can achieve extremely high data rates (exceeding 1~Gbit/s in laboratory conditions) but require strict line-of-sight and are quickly attenuated by water~\cite{saeed2019underwater, wubben2020challenges}. Tests show that clear ocean water allows ranges from 60 to 70~m, while turbid water reduces it below 10~m. Optical links consume far less energy per bit than acoustics, making them attractive for short-range high-data-rate tasks such as docking or sensor payload transmission. RF communication in water is generally confined to very shallow or short-range use due to seawater conductivity, so RF is typically limited to intra-vehicle links or surface-to-underwater communication~\cite{wubben2020challenges}.

Magnetic Induction (MI) communication offers another short-range alternative, using time-varying magnetic fields with negligible propagation delay and stable channel characteristics~\cite{lodovisi2018performance}. Although limited to short distances (under 2~m), MI channels are promising for local coordination due to their low sensitivity to water~properties.

To balance these trade-offs, hybrid acoustic--optical--MI networks have been proposed~\cite{wubben2020challenges, lodovisi2018performance}. In such multi-modal schemes, long-range acoustic links maintain connectivity while optical or MI links provide high-speed local exchange. Adaptive switching protocols select appropriate channels based on environmental conditions and data priority~\cite{li2025recent}. Control data requiring robustness may be routed acoustically or magnetically, while sensor data streams can exploit optical channels when available. These architectures aim to deliver resilience and flexibility for mission-critical swarm behaviour. For bio-inspired underwater swarms, these physical communication modalities define the bandwidth, latency, and 
range constraints under which biologically inspired coordination algorithms (such as AFSA, WOA, CRO, and 
MPA) and signalling schemes must operate, directly shaping feasible neighbourhood sizes, update rates, and 
degrees of~decentralisation.

In the classification framework introduced in Section~\ref{framework}, these choices of physical communication modality and hybrid architecture primarily influence the communication-dependency and energy-efficiency dimensions by constraining how often and how far agents can exchange~information.

\subsection{Network Topologies for Swarm~Connectivity}

Underwater swarm networks must operate under high latency, intermittent connectivity, and 
dynamic node positioning. Conventional terrestrial networking paradigms are not directly applicable due to the slow speed of sound and time-varying link quality~\cite{pal2022communication, theocharidis2025underwater}. Adaptive and opportunistic routing techniques have thus been developed for underwater~conditions.

Opportunistic routing protocols utilise the broadcast nature of acoustic communication, allowing multiple candidate forwarders to compete and cooperate in relaying messages~\cite{wubben2020challenges}. This avoids fixed paths and supports dynamic adaptation. Geographic routing methods transmit data towards destination coordinates, assuming nodes have relative positioning capabilities. In three-dimensional underwater environments, depth-based routing can be effective, with 
vertical position as the primary metric~\cite{li2022adaptive}. Adaptive multi-zone protocols integrate geographic and depth-aware strategies, selecting routes based on node density and link~quality.

\textls[-7]{TDMA (Time-Division Multiple Access) schemes support collision-free communication by allocating time slots for each node. Despite the challenges of long propagation delays and clock drift, TDMA has been successfully implemented using guard intervals and adaptive synchronisation to maintain reliable timing, as 
demonstrated in field deployments~\cite{gussen2021optimisation}.}

Swarm topologies vary depending on mission goals. Linear chains are effective for pipeline inspection or relay tasks, while clustered or mesh topologies distribute connectivity for broader coverage. Decentralised networks allow robustness against node failure by distributing control and routing responsibilities, while hybrid schemes combine hierarchical control and local autonomy~\cite{pal2022communication}. The optimal topology is mission-specific and must accommodate density, mobility, and 
environmental disruption, emphasising the importance of adaptive communication architectures. From a bio-inspired perspective, these network topologies underpin how local interaction rules are realised at system level, constraining how quickly bio-inspired behaviours such as schooling, foraging, and 
predator--prey dynamics can propagate through the swarm under realistic acoustic and optical~connectivity.

\subsection{Bio-Inspired Communication~Protocols}

Bio-inspired communication protocols take cues from nature, where organisms coordinate efficiently under severe sensory and communication limitations. Such protocols are inherently suited to underwater conditions, as 
they minimise bandwidth use while maintaining functional coordination~\cite{liu2013covert, zhao2025bioinspired}.

Pheromone-inspired strategies enable indirect communication through persistent environmental markers, analogous to chemical trails in ant colonies~\cite{geng2019pheromone, li2019pheromone}. Underwater, virtual markers and temporary data gradients can represent shared information, allowing agents to influence each other’s trajectories without direct message exchange. This indirect form of communication supports cooperative exploration and foraging while conserving~bandwidth.

Acoustic mimicry provides another biologically inspired technique. By emulating the spectral patterns of marine mammal vocalisations, systems can achieve covert and environmentally integrated communication~\cite{liu2013covert}. Such methods may reduce interference and detection while blending with natural acoustic backgrounds, although 
they trade efficiency for~concealment.

Collective sensing constitutes an implicit communication approach where agents infer swarm state from neighbour movement and environmental cues rather than explicit messages. For example, AFSA’s visual perception model enables coordination through local observation of neighbours, analogous to visual alignment in fish schools~\cite{pourpanah2022afsa}. These methods reduce dependence on message exchange and improve resilience to packet~loss.

Bio-inspired protocols emphasise efficiency and robustness. When integrated with swarm coordination algorithms, they enable communication-constrained systems to maintain coherent group behaviour using minimal shared information, an 
essential property for scalable underwater~operations.

\subsection{Delay Tolerance, Fault Handling, and Synchronisation}
The underwater communication channel imposes significant propagation delays, intermittent connectivity, and 
link unreliability, making delay-tolerant networking essential for swarm operation~\cite{wubben2020challenges}. Rather than requiring continuous end-to-end connections, Delay-Tolerant Networks (DTNs) allow nodes to store, carry, and 
forward information when connectivity becomes available. These mechanisms enable effective cooperation even under sporadic contact~conditions.

Acoustic propagation delays, often several seconds over kilometre-scale ranges, disrupt real-time coordination and synchronisation~\cite{gussen2021optimisation}. Delay-tolerant coordination leverages prediction and local autonomy to maintain functionality despite outdated information. Algorithms such as WOA and MPA, which depend primarily on local states and global trends rather than direct communication, naturally accommodate these delays~\cite{mirjalili2016woa, faramarzi2020mpa}.

Fault handling ensures continued swarm function despite individual agent and link failures. Decentralised designs inherently improve fault tolerance, as 
no single unit is mission-critical. In CRO-inspired frameworks, competitive depredation mechanisms naturally replace failed and underperforming agents with new participants~\cite{salcedo2014cro}. Redundant routing paths and adaptive link selection further enhance~reliability.

Synchronisation remains a challenge due to clock drift and long propagation delays. TDMA-based coordination schemes require guard intervals and periodic re-synchronisation. Adaptive synchronisation adjusts transmission timing based on measured delay, while asynchronous coordination eliminates dependency on shared time references by relying on local interaction cues and environmental feedback. These strategies collectively enable stable swarm coordination under uncertain temporal~conditions.

Environmental variability compounds these challenges by altering sound-speed profiles, attenuation, and 
noise levels. Adaptive modulation, dynamic power control, and 
predictive environment modelling can mitigate performance degradation by preemptively adjusting communication parameters to maintain connectivity and~throughput.

\subsection{Cross-Layer Optimisation for Cooperative~Swarms}
Cross-layer optimisation integrates communication, control, and 
energy management to improve overall swarm performance rather than optimising each subsystem in isolation~\cite{gussen2021optimisation, khan2025hybrid}. This holistic approach is particularly valuable in underwater robotics, where communication cost and reliability strongly influence system design and coordination~strategies.

Communication is often the most energy-intensive process in an underwater robot, with 
transmission power dominating total energy consumption~\cite{khan2025hybrid}. Cross-layer optimisation therefore focuses on balancing communication reliability with energy efficiency. Algorithms such as WOA and MPA, characterised by low communication dependency, are naturally suited to energy-aware operation~\cite{mirjalili2016woa, faramarzi2020mpa}. AFSA and CRO, which involve frequent neighbour interactions, require additional energy management strategies for long-duration missions~\cite{pourpanah2022afsa, salcedo2014cro}.

Adaptive physical-layer techniques adjust OFDM subcarrier allocation and modulation parameters based on environment feedback to optimise BER and energy use~\cite{gussen2021optimisation}. These adjustments enable swarms to maintain data integrity while reducing redundant transmission energy costs. Energy-efficient communication protocols can reduce transmission power from approximately 30 to 50\% through adaptive modulation and dynamic power control, directly extending mission endurance in energy-limited swarm deployments~\cite{gussen2021optimisation}.

Hybrid communication architectures further enhance cross-layer optimisation. By switching intelligently between acoustic, optical, and 
MI modalities, systems can allocate bandwidth and energy according to task urgency and environmental suitability~\cite{lodovisi2018performance}. Low-rate, high-reliability control data may use acoustic and magnetic channels, while high-volume sensing data employ optical links where visibility allows. This dynamic selection minimises energy consumption while sustaining mission~performance.

Cross-layer integration creates synergy between communication protocols and coordination algorithms. Algorithms that exploit local sensing and indirect communication reduce network traffic, while communication strategies that support such decentralised coordination reinforce autonomy and robustness. Hardware platforms with multimodal communication interfaces enable these adaptive behaviours, allowing swarms to reconfigure communication and control strategies dynamically in response to mission and environmental~demands.

Cross-layer optimisation thus represents a critical enabler of practical underwater swarm robotics. By jointly considering communication constraints, algorithmic design, and 
hardware capabilities, it is possible to achieve efficient, resilient, and 
scalable cooperation that meets mission objectives within real-world~limitations.

\section{System Design and~Implementation}
\label{systemdesign}

Effective underwater swarm robotics requires careful integration of hardware platforms, sensor systems, energy management, and 
software architectures to enable reliable operation in challenging marine environments. This section examines the principal design considerations for implementing underwater swarm systems, from 
hardware architectures and sensor integration to energy optimisation and experimental validation. The discussion covers both theoretical design principles and practical implementation experiences from deployed systems, providing insight into the trade-offs and constraints that shape real-world underwater swarm deployments. In the context of this review, these design choices are interpreted through a bio-inspired lens, highlighting how platform, sensing, and 
power decisions support or constrain biologically motivated coordination and communication~strategies.

\subsection{Hardware Architectures for Underwater~Swarms}

Underwater swarm platforms must balance trade-offs among vehicle size, endurance, communication capability, autonomy, and 
cost. Two benchmark implementations that have validated decentralised coordination, TDMA-based scheduling, and 
adaptive physical-layer strategies in real underwater environments are the COMET and NEMOSENS projects~\cite{gussen2021optimisation}. The COMET AUV (approximately 2~m in length, 40~kg) provides extended operational duration (20 h) and deep-sea capacity (300~m), featuring high-endurance acoustic communication and INS/DVL navigation systems. However, its larger size limits scalability for large swarm formations. In contrast, NemoSens (0.9~m, 9~kg) is a compact swarm-ready platform with moderate endurance (10~h), flexible modularity for communication and navigation, and 
Linux-based modular control. Both platforms have demonstrated effective decentralised coordination where each AUV maintains network awareness and communication order without central nodes, with 
TDMA successfully implemented using guard intervals and adaptive synchronisation to maintain reliable timing despite long propagation delays and clock drift. Adaptive physical-layer techniques demonstrated in these projects adjust OFDM subcarrier allocation and modulation parameters based on environment feedback to optimise BER and energy use, enabling swarms to maintain data integrity while reducing redundant transmission energy costs. These projects followed a progressive validation pipeline from simulation through laboratory testing to field verification, providing critical empirical validation of swarm coordination~strategies.

Explicitly designed for swarm coordination, MONSUN~II offers a small (60~cm), low-cost (4.2~kg) architecture integrating six thrusters for 5-DOF mobility, onboard sensors (IMU, depth, temperature, IR), and 
modular I\textsuperscript{2}C/SPI expansion ports for scalable integration~\cite{osterloh2012monsun}. Its open hardware design simplifies maintenance and supports plug-and-play experimentation. These architectures prioritise robustness and rapid deployment in shallow~environments.

Bio-inspired miniature designs such as Bluebot enable vision-based schooling behaviour using LED-based communication and fin-like actuation for precise 3D motion in confined water columns~\cite{berlinger2021bluebot}. Platforms like M-AUE take advantage of buoyancy-driven drifting with vertical mobility for submesoscale ocean dynamics tracking, providing high spatial resolution data via collective deployment~\cite{jaffe2017swarm}. Open-source MURs such as the one proposed by Mayberry et~al.~\cite{Mayberry2025} incorporate 5-DOF propulsion, multiple cameras, and 
ROS-native controllers for rapid~experimentation.

Table~\ref{tab:swarm_platforms} compares representative swarm-capable underwater robots based on structural and functional~parameters.

\begin{table}[H]
\caption{Representative underwater swarm-capable robotic platforms and their~characteristics.}
\label{tab:swarm_platforms}
\small

\begin{adjustwidth}{-\extralength}{0cm}
\begin{tabularx}{\linewidth}{lccccm{7cm}<{\raggedright}}
\toprule
\textbf{Platform} & \textbf{Length} & \textbf{Weight} & \textbf{Endurance} & \textbf{Max Depth} & \textbf{Notes} \\
\midrule
COMET AUV~\cite{gussen2021optimisation} & 2.0~m & <40~kg & $\sim$20~h & 300~m & High-endurance, acoustic comms, INS/DVL~navigation \\
\midrule
NemoSens~\cite{gussen2021optimisation} & 0.9~m & <9~kg & $\sim$10~h & 300~m & Compact swarm-ready platform with Linux-based modular~control \\
\midrule
MONSUN II~\cite{osterloh2012monsun} & 0.6~m & 4.2~kg & $\sim$10~h & 10~m & Six-thruster design, modular I\textsuperscript{2}C/SPI architecture for swarm~testing \\
\midrule
Bluebot (BlueSwarm)~\cite{berlinger2021bluebot} & 0.13~m & \hl{N/A} 
 & Lab-scale & Shallow & Bio-inspired vision-based schooling, fin actuation, 3D~motion \\
\midrule
M-AUE~\cite{jaffe2017swarm} & 0.3~m & N/A & $\sim$8~h & 100~m & Miniature drifting robots for subsurface layer~tracking \\
\midrule
Open-source MUR~\cite{Mayberry2025} & 0.4~m & N/A & N/A & 5 to 10~m & 5-DOF propulsion, ROS, WiFi, camera suite; designed for modular swarm~use \\
\bottomrule
\end{tabularx}
\end{adjustwidth}

\noindent{\footnotesize{%
\hl{Sources}: 
Gussen et~al. (2021)~\cite{gussen2021optimisation}, Osterloh et~al. (2012)~\cite{osterloh2012monsun}, Berlinger et~al. (2021)~\cite{berlinger2021bluebot}, Jaffe et~al. (2017)~\cite{jaffe2017swarm}, Mayberry et~al. (2025)~\cite{Mayberry2025}, Chen et~al. (2025)~\cite{s25206413}.
}}
\end{table}

Distributed onboard compute is central to swarm behaviour, enabling decentralised formation control and adaptive mission execution. For instance, swarm platforms have demonstrated effective formation behaviour with each AUV running an ROS-based controller on Raspberry Pi 4 boards~\cite{s25206413}. Architectures emphasise modular sensor integration, hot-swappable interfaces, and 
energy-efficient propulsion. Communication systems range from acoustic modems for deep water to WiFi and optical signalling in confined or shallow domains~\cite{zhang2023omnidirectional}.

Together, these hardware architectures span a design spectrum ranging from robust long-endurance AUVs to bio-inspired miniature swarming agents, offering a broad toolkit for scalable underwater multi-agent systems. For bio-inspired underwater swarms, selecting along this spectrum determines how closely physical platforms can emulate the sensing, manoeuvrability, and 
interaction scales assumed by biological analogues such as fish schools or drifting organisms, and, 
thus, how faithfully bio-inspired coordination rules can be implemented in~practice.

\textls[-6]{These platform decisions most directly affect swarm scalability and environmental adaptability, setting practical bounds on the number of vehicles that can be deployed and the range of ocean conditions under which bio-inspired behaviours can be reliably~sustained.}

\subsection{Sensor Integration and Environmental~Perception}
Sensor integration provides the perceptual foundation for navigation, obstacle avoidance, and 
environmental monitoring in underwater swarms. Acoustic sensors such as sonar deliver long-range detection unaffected by turbidity, while optical sensors offer high-resolution imaging but are constrained by light attenuation and scattering~\cite{cong2021underwater, jiang2024low}. 

\textls[-5]{Sensor fusion frameworks combine acoustic, optical, magnetic, and 
bio-inspired modalities to improve reliability in heterogeneous conditions~\cite{cong2021underwater}. Navigation under GPS denial relies on INS, dead reckoning, and 
acoustic localisation systems~\cite{sorensen2025localization}. INS-DVL integration remains a standard method for maintaining positional accuracy between acoustic~updates.}

Environmental perception also supports implicit coordination through local sensing. AFSA’s visual perception model exemplifies this approach, where agents infer collective behaviour \textls[-28]{by observing neighbours’ movement patterns~\cite{pourpanah2022afsa}. This perception-based coordination reduces message exchange, which is particularly beneficial under acoustic bandwidth~constraints.}

\begin{figure}[H]\hspace{-1.3mm}
\includegraphics[width=0.6\linewidth]{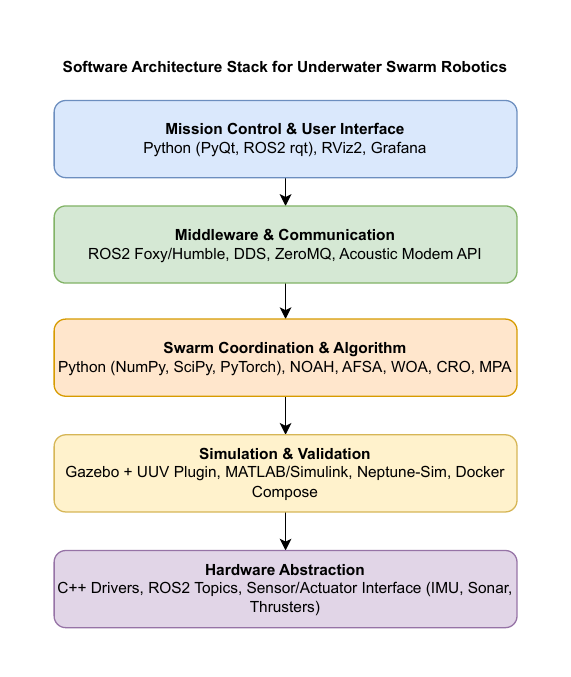}
\caption{\hl{Software} 
 architecture for underwater swarm robotics systems, illustrating the integration of distributed control, communication protocols, sensor fusion, and mission planning modules within a cohesive multi-agent platform~architecture.}
\label{fig:software_architecture}
\end{figure}

Integrating communication hardware requires balancing range, bandwidth, and 
power consumption. Acoustic modems provide long-range links with high latency; optical systems deliver high-speed, short-range data exchange; and hybrid configurations combine both for adaptive performance~\cite{gussen2021optimisation, wubben2020challenges}. Payload capacity limits dictate sensor selection, necessitating modular architectures that allow mission-specific reconfiguration while maintaining overall hydrodynamic efficiency. Figure~\ref{fig:software_architecture} illustrates the software architecture for underwater swarm robotics systems, showing the integration of distributed control, communication protocols, sensor fusion, and 
mission planning~modules.
\vspace{-6pt}

\subsection{Energy Management and Power~Optimisation}
Energy management defines operational duration and mission feasibility for underwater swarms~\cite{guo2019sustainable, zhang2023autonomous}. Communication dominates total energy use, requiring optimisation of transmission power, duty cycling, and 
coordination frequency~\cite{khan2025hybrid, gussen2021optimisation}. 

\textls[-5]{Lithium-ion batteries remain the preferred energy source, balancing density, power output, and 
safety. Field deployments have demonstrated operational endurance ranging from 10 to 20~h under typical operational loads~\cite{gussen2021optimisation}. Emerging hydrogen fuel cell systems and energy harvesting from ocean currents represent potential extensions to future~endurance.}

Biological collectives inspire energy-efficient coordination. Bio-inspired algorithms such as WOA and MPA minimise communication overhead and motion redundancy to lower energy demand~\cite{mirjalili2016woa, faramarzi2020mpa}. AFSA and CRO require tighter energy budgeting due to neighbourhood evaluation processes~\cite{pourpanah2022afsa, salcedo2014cro}. Hydrodynamic interactions and ocean current direction strongly influence power requirements during obstacle avoidance manoeuvres, underscoring the importance of strategic trajectory planning~\cite{cai2021stream}.

Propulsion design has a significant influence on energy use. Conventional thrusters are reliable but less efficient at low speeds; bio-inspired mechanisms such as undulating fins and oscillating foils offer enhanced efficiency and stealth~\cite{cong2021underwater}. Recent advances in miniature underwater robots have explored diverse actuation methods including motors, magnetic field actuation, piezoelectric actuators, and 
soft materials (e.g., shape memory alloys, dielectric elastomer actuators, ionic polymer-metal composites), each offering distinct trade-offs in power consumption, response speed, and 
motion flexibility~\cite{wang2025actuation}. These actuation technologies enable various locomotion modes such as fish-inspired swimming, jetting, paddling, and 
crawling, providing flexibility for different swarm applications and environmental conditions~\cite{wang2025actuation}. Energy-aware path planning further improves endurance by exploiting ocean currents for passive drift and current-assisted travel~\cite{cai2021stream}. Empirical studies indicate that energy-aware path planning can reduce propulsion energy consumption by approximately 20 to 30\% through strategic current exploitation and trajectory optimisation~\cite{cai2021stream}. Cross-layer power management, which coordinates propulsion, sensing, and 
communication demands, ensures optimal use of limited energy reserves throughout a~mission.

Within the classification framework, these energy management strategies directly shape the energy-efficiency dimension and indirectly support scalability by determining how mission duration and swarm size can be traded off against communication and propulsion~costs.

\subsection{Simulation and Experimental~Platforms}
Simulation and testing environments enable validation of algorithms, communication models, and 
control architectures before costly field trials. Effective underwater simulation must capture hydrodynamic resistance, buoyancy variation, acoustic propagation delay, and 
environmental noise~\cite{ciuccoli2024underwater, szlkeg2022simulation}. Realistic three-dimensional modelling enhances predictive fidelity for formation control and coordination algorithms~\cite{kaur2024energy}.

Robotics Software Frameworks (RSFs) and Multi-Agent System Frameworks (MASFs) provide essential middleware for distributed simulation and control~\cite{inigo2012robotics}. These platforms abstract hardware dependencies, support inter-agent communication, and 
enable scalable swarm testing. Middleware must manage limited bandwidth, latency, and 
service discovery within dynamic underwater topologies~\cite{wubben2020challenges}. In this review, the 
focus remains on these functional capabilities rather than on specific software stacks, to 
avoid unnecessary architectural detail and potential confusion for readers primarily interested in bio-inspired coordination, communication, and 
system~design.

Validation through simulation-experiment integration is critical. A progressive validation pipeline from simulation through laboratory testing to field verification has been demonstrated in successful deployments~\cite{gussen2021optimisation}. Laboratory environments (tank facilities, acoustic testbeds, and 
optical calibration setups) offer controlled conditions to evaluate sensor reliability and subsystem performance before open-water deployment. Hardware-in-the-Loop (HIL) simulation bridges software and hardware domains, providing realistic testing of sensors, communications, and 
control algorithms under repeatable conditions~\cite{cong2021underwater}. This multi-stage validation reduces development risk and improves system robustness. Table~\ref{tab:validation_framework} presents a progressive validation framework for underwater swarm robotics testing, illustrating the structured pathway from simulation through HIL testing to controlled field trials. These simulation and validation frameworks provide the testbed in which bio-inspired swarm algorithms and communication protocols can be exercised under controlled yet realistic conditions, enabling systematic assessment of how closely biologically motivated behaviours translate into reliable underwater~performance.

\begin{table}[H]
\caption{Progressive validation framework for underwater swarm robotics testing. 
The framework illustrates a structured pathway from simulation through Hardware-in-the-Loop testing to controlled field trials.}
\label{tab:validation_framework}
\footnotesize

\begin{adjustwidth}{-\extralength}{0cm}
\setlength{\tabcolsep}{1.8mm}
\begin{tabularx}{\linewidth}{%
m{2.5cm}<{\raggedright}%
m{3.5cm}<{\raggedright}%
m{3.7cm}<{\raggedright}%
m{3.5cm}<{\raggedright}%
m{3.5cm}<{\raggedright}
}
\toprule
\textbf{Validation Level} & \textbf{Test Environment} & \textbf{Protocols} & \textbf{Success Criteria} & \textbf{Data Collection} \\
\midrule
Level 1: Algorithm Validation & Simulation platforms (e.g.,~Gazebo, MATLAB/Simulink, DESERT Underwater) & Monte Carlo runs, synthetic datasets, parameter sweeps, comparative analysis & Algorithm convergence, computational efficiency, cross-run consistency & Algorithm metrics, convergence data, computation time \\
\midrule
Level 2: Hardware-in-the-Loop & Water-tank testing, acoustic-modem integration & Sensor integration, communication-delay characterisation, energy~profiling & Functional integration, link reliability, sensor accuracy & Sensor data, communication logs, energy profiles \\
\midrule
Level 3: Laboratory Testing & Controlled pool/tank environment, multi-AUV deployment & Coordination and avoidance validation, formation control~tests & Collision-free operation, formation maintenance, swarm coherence & Position tracking, formation metrics, coordination logs \\
\midrule
Level 4: Field Trials & Open-water marine environment & Extended missions, environmental adaptation, system-resilience evaluation & Mission success, operational reliability, environmental robustness & Environmental data, mission logs, performance~metrics \\
\bottomrule
\end{tabularx}
\end{adjustwidth}

\noindent{\footnotesize{%
\hl{Sources:} Adapted from Zhao et~al.~\cite{zhao2025bioinspired}, Connor et~al.~\cite{connor2021current}, Cong et~al.~\cite{cong2021underwater}, Cai et~al.~\cite{cai2023cooperative}.
}}
\end{table}

\subsection{Emerging Prototypes and Field~Deployments}
Despite increasing simulation maturity, few swarm systems have reached sustained field operation~\cite{connor2021current}. Benchmark implementations have validated decentralised coordination, TDMA-based scheduling, and 
adaptive physical-layer strategies in real underwater environments~\cite{gussen2021optimisation}. 

Field testing requires strict operational and safety protocols, encompassing environmental monitoring (temperature, salinity, turbidity, current, and 
noise), real-time telemetry, and 
recovery systems. Mission evaluation metrics include communication reliability, energy consumption, localisation accuracy, and 
task success rate. Safety procedures cover emergency recovery, fail-safe shutdown, and 
redundancy protocols for lost communication~links.

Operational scenarios demonstrate practical relevance: coral reef monitoring, sub-sea pipeline inspection, search and rescue, and 
deep-sea exploration~\cite{cai2023cooperative, zhao2025bioinspired}. Each scenario tests algorithms under distinct environmental constraints, from 
shallow high-turbidity zones to high-pressure deep-sea conditions. Incremental field validation (simulation to laboratory to real environment) enables systematic fault isolation and progressive risk reduction~\cite{zhao2025bioinspired, gussen2021optimisation}.

\subsection{Metrics for Evaluating Swarm~Performance}

Evaluating underwater swarm performance requires multidimensional metrics covering coordination, communication, energy efficiency, robustness, and 
scalability. Coordination is typically assessed through velocity correlation, nearest-neighbour distance, and 
coverage yield. Navigation accuracy is often measured using cross-track error and formation deviation~\cite{zhu2023multiAUVHRL}. Communication performance is quantified by Packet Delivery Ratio (PDR), end-to-end latency, and 
throughput. For instance, the 
EVA framework achieved over 96\% PDR in large-scale UWSN simulations~\cite{jmse9111219}, while the ILAF protocol demonstrated 99\% delivery reliability~\cite{ali2024ilaf}. AQUA-GLOMO simulations recorded control delays between 0.1--0.54~s~\cite{dhurandher2012aquaglomo}. 

Energy metrics such as joules per metre and energy-to-completion ratios indicate mission efficiency. Hierarchical reinforcement learning-based navigation reduced formation error to below 0.7~m~\cite{zhu2023multiAUVHRL}, and 
DSUA-based cooperative control reduced energy use by 12.6\%~\cite{zhang2024dsua}. Fault-tolerant control schemes maintain swarm coordination with under 15\% performance degradation~\cite{chen2025ftmpc}. Scalability is validated by maintaining over 80\% baseline performance at 400+ nodes~\cite{jmse9111219}. Representative benchmark thresholds commonly reported in the literature and used for underwater swarm validation are summarised in Table~\ref{tab:performance_thresholds}.

\begin{table}[H]
\caption{\hl{Representative} 
 performance benchmarks for underwater swarm~validation.}
\label{tab:performance_thresholds}
\small
\begin{adjustwidth}{-\extralength}{0cm}
\setlength{\tabcolsep}{4.8mm}
\begin{tabularx}{\linewidth}{@{\hspace{2mm}}lcl}
\toprule
\textbf{Metric} & \textbf{Target Threshold} & \textbf{Reference/Justification} \\
\midrule
Packet Delivery Ratio (PDR) & $\geq$95\% & EVA: 96\% at 400 nodes~\cite{jmse9111219}; ILAF: 99\% reliability~\cite{ali2024ilaf} \\
Formation Error (m) & $\leq$0.5 & RL-based multi-AUV: <0.7~m error~\cite{zhu2023multiAUVHRL} \\
Energy Efficiency (J/m) & $\leq$1.2$\times$ single-AUV baseline & DSUA navigation: 12.6\% energy reduction~\cite{zhang2024dsua} \\
Latency (s) & $\leq$2~s & AQUA-GLOMO: 0.1--0.54~s delay~\cite{dhurandher2012aquaglomo} \\
Fault Tolerance Degradation & $\leq$15\% & MPC control: loss $\leq$15\% under failure~\cite{chen2025ftmpc} \\
Scalability Performance Ratio & $\geq$0.8 & EVA: Maintained 96\% PDR at 400 nodes~\cite{jmse9111219} \\
\bottomrule
\end{tabularx}
\end{adjustwidth}

\noindent{\footnotesize{%
\hl{Sources:} Adapted from Khasawneh et~al.~\cite{jmse9111219}, Ali et~al.~\cite{ali2024ilaf}, Zhu et~al.~\cite{zhu2023multiAUVHRL}, Zhang et~al.~\cite{zhang2024dsua}, Dhurandher et~al.~\cite{dhurandher2012aquaglomo}, and 
Chen et~al.~\cite{chen2025ftmpc}.
\hl{Note:} 
Thresholds represent values derived from empirical studies and serve as benchmarks for system validation and~comparison.
}}

\end{table}

\section{Classification and Synthesis~Framework}
\label{framework}

The preceding sections have examined bio-inspired coordination mechanisms, communication strategies, and 
system design considerations as distinct domains. However, effective underwater swarm robotics requires systematic integration of these components to address the complex interdependencies that shape real-world system performance. This section presents a multi-dimensional classification framework that synthesises findings across algorithms, communication, and 
hardware domains, enabling evaluation and selection of coordination strategies based on mission requirements and operational~constraints.

\subsection{Dimensions of Classification: Communication, Adaptability, Energy, and Scalability}

Effective classification of bio-inspired algorithms for underwater swarm robotics requires evaluation across multiple dimensions that reflect the fundamental constraints and requirements of marine environments. Drawing on characteristics identified in the underwater swarm robotics literature~\cite{zhao2025bioinspired}, this framework evaluates algorithms across four critical dimensions: communication dependency, environmental adaptability, energy efficiency, and 
swarm scalability. Conceptually, the 
resulting four-dimensional classification space can be visualised as a multi-dimensional matrix in which each algorithm occupies a distinct region according to its characteristics in communication dependency, environmental adaptability, energy efficiency, and 
swarm~scalability.

\hl{Communication dependency} 
measures the extent to which an algorithm relies on explicit message exchange or global state information for effective coordination. Algorithms with low communication dependency, such as WOA and MPA, operate effectively with minimal message exchange, making them well suited to acoustic environments characterised by bandwidth limitations and propagation delays~\cite{mirjalili2016woa, faramarzi2020mpa}. Algorithms with moderate communication dependency, such as AFSA and CRO, require neighbourhood interactions or local information exchange, necessitating careful bandwidth management~\cite{pourpanah2022afsa, salcedo2014cro}. This dimension directly influences algorithm selection based on available communication infrastructure and energy~constraints.

\hl{Environmental adaptability} captures an algorithm's ability to maintain performance under varying marine conditions, including currents, temperature gradients, turbidity, and 
pressure variations. Algorithms inspired by marine organisms that naturally adapt to environmental change, such as AFSA's fish schooling behaviours and MPA's predator--prey dynamics, demonstrate high adaptability~\cite{adeli2012review, mccluskey2021foraging}. WOA's bubble-net hunting strategies adapt to prey motion and ocean currents, while CRO's competitive mechanisms respond to environmental changes through reef-space dynamics~\cite{clapham2000humpback, garciahernandez2020cro}. This dimension is critical for missions operating across diverse environmental~conditions.

\hl{Energy efficiency} evaluates computational complexity and communication overhead, both of which affect mission duration and operational feasibility. All four marine-specific algorithms operate at $O(N \cdot D)$ complexity per iteration, as 
reported for WOA~\cite{mirjalili2016woa}, MPA~\cite{faramarzi2020mpa}, AFSA~\cite{pourpanah2022afsa}, and 
CRO~\cite{salcedo2014cro}, but 
they differ in parameter requirements and neighbourhood computation. WOA and MPA offer efficient operation with minimal control parameters, whereas AFSA and CRO incur additional overhead due to neighbourhood evaluation and settlement processes. Communication energy costs further differentiate algorithms, with 
low-communication approaches naturally supporting extended mission~duration.

\hl{Swarm scalability} assesses algorithm performance as the number of agents increases, evaluating whether coordination mechanisms remain effective across different swarm sizes. Algorithms that rely primarily on local interactions, such as AFSA's neighbourhood-based coordination, scale well through distributed mechanisms~\cite{pourpanah2022afsa}. WOA and MPA also demonstrate good scalability through population-based optimisation that scales linearly with swarm size~\cite{alyasseri2024woa, makhadmeh2025mowoa}. CRO's grid-based structure provides moderate scalability, with 
performance dependent on grid size and reproduction parameters~\cite{salcedo2014cro}.

These four dimensions are not independent; they interact in ways that create complex trade-offs. Algorithms with low communication dependency often exhibit higher energy efficiency but may sacrifice coordination fidelity. High environmental adaptability may demand greater computational resources, reducing energy efficiency. Understanding these interdependencies enables systematic algorithm selection based on mission-specific priorities and~constraints.

\subsection{Mapping of Existing Bio-Inspired~Algorithms}

The classification framework enables systematic mapping of marine-specific bio-inspired algorithms across the four-dimensional space, revealing distinct positioning that reflects their biological inspiration and computational characteristics. This mapping provides a structured basis for understanding algorithm suitability for different mission requirements and operational constraints. Table~\ref{tab:comprehensive_comparison} presents a comprehensive comparison of marine-specific bio-inspired algorithms for underwater swarm robotics~applications.

\startlandscape
\begin{table}[H]
\caption{Comprehensive comparison of marine-specific bio-inspired algorithms for underwater swarm robotics~applications.}
\label{tab:comprehensive_comparison}
\footnotesize
\setlength{\tabcolsep}{1.9mm}
\begin{tabularx}{\linewidth}{l
m{2.6cm}<{\raggedright}
m{2.9cm}<{\raggedright}c
m{1.7cm}<{\centering}
m{2.2cm}<{\raggedright}
m{2.3cm}<{\raggedright}
m{2.2cm}<{\raggedright}
m{2.8cm}<{\raggedright}
m{3.8cm}<{\raggedright}}
\toprule
\textbf{Algorithm} & \textbf{Inspiration} & \textbf{Parameters} & \textbf{Complexity} & \textbf{Typical Pop.} & \textbf{Comm. Pattern} & \textbf{Efficiency} & \textbf{Scalability} & \textbf{Convergence} & \textbf{Applications} \\
\midrule
AFSA & Fish preying, swarming, following behaviours & Visual ($v$), Step ($s$), try\_number, crowding ($\delta$); $N$, $D$ 
 & $O(N \cdot D)$ & 50 to 100 & Moderate neighbourhood interaction & Simple rules; distance computation overhead & Linear in $N$; diversity maintained via crowding factor & Balanced exploration/exploitation; slower convergence & UWSN routing and topology; multi-AUV coordination; acoustic constrained navigation; hybrid AFSA~variants \\
\midrule
WOA & Humpback whale bubble net hunting & Decay factor $a$ (2 to 0), spiral const $b$, probability $p = 0.5$; $N$, $D$ & $O(N \cdot D)$ & 30 to 50 & Low; global best-oriented updates & Few parameters $(a, b, p)$; minimal overhead & Linear in $N$; inherently parallelisable & Strong exploitation via spiral; fast convergence & AUV 3D path planning; multi-AUV formation; obstacle~avoidance; current-aware navigation; energy-efficient~routing \\
\midrule
CRO & Coral broadcast spawning, brooding, larvae settling, depredation & Reef grid ($R \times C$), broadcast fraction $F_b$, depredation prob $P_d$, settle attempts $\kappa$; $N$, $D$ & $O(N \cdot D)$ & 50 to 100 (grid-based) & Moderate; reef wide spatial competition & Settlement bookkeeping; comparable to EA & Linear in $N$; sensitive to grid size and $F_b$, $P_d$ & Balanced via spawning/brooding; fight for space controls exploitation & Potential: underwater infrastructure placement; current validation limited to general optimisation~problems \\
\midrule
MPA & Marine predator Lévy and Brownian movement; FAD~effects & Phase-based (Brownian/Lévy ratio); FAD probability; velocity ratio; $N$, $D$ & $O(N \cdot D)$ & 10 to 50 & Low; best solution and phase driven updates & Derivative-free; minimal tuning; efficient foraging model & Linear in $N$; phase transitions handle scale & Fast exploration (Lévy); local refinement (Brownian); FAD avoids~stagnation & UWSN deployment and coverage; multiobjective AUV task allocation; adaptive marine sampling; swarm~coordination \\
\bottomrule
\end{tabularx}

\begin{adjustwidth}{+\extralength}{0cm}
\noindent{\footnotesize{\hl{Sources}: 
Adeli (2012)~\cite{adeli2012review}, Pourpanah et~al. (2022)~\cite{pourpanah2022afsa}, Mirjalili and Lewis (2016)~\cite{mirjalili2016woa}, Nadimi-Shahraki et~al. (2023)~\cite{nadimiShahraki2023woa}, Salcedo-Sanz et~al. (2014)~\cite{salcedo2014cro}, Faramarzi et~al. (2020)~\cite{faramarzi2020mpa}, Al-Betar et~al. (2023)~\cite{albetar2023mpa}, Chun et~al. (2024)~\cite{chun2024improved}, Bujok (2023)~\cite{bujok2023evaluation}. Population sizes represent typical ranges reported across benchmark studies; actual implementations vary by problem dimension and resource~constraints.}}
\end{adjustwidth}

\end{table}

\finishlandscape

AFSA occupies a position characterised by moderate communication dependency, high environmental adaptability, moderate energy efficiency, and 
good scalability~\cite{pourpanah2022afsa, adeli2012review}. AFSA's fish-schooling inspiration provides natural adaptation to marine conditions through collective behaviour, while its visual perception system enables coordination through environmental sensing rather than explicit message passing. The algorithm's moderate communication requirements stem from neighbourhood interactions needed for swarming and following behaviours, creating a balance between coordination effectiveness and communication~efficiency.

WOA demonstrates low communication dependency, high environmental adaptability, high energy efficiency, and 
good scalability. Mirjalili and Lewis introduced the Whale Optimisation Algorithm~\cite{mirjalili2016woa}, Rana et~al. reviewed subsequent applications and modifications~\cite{rana2020whale}, and 
Yan et~al. demonstrated an AUV 3D path-planning formulation based on WOA~\cite{yan2022woa}. WOA's bubble-net hunting strategy requires minimal explicit communication, with 
coordination emerging from shared environmental cues and global best solutions. The algorithm's streamlined optimisation process with few control parameters contributes to high energy efficiency, while its population-based structure enables effective scaling with swarm~size.

CRO exhibits moderate communication dependency, moderate environmental adaptability, moderate energy efficiency, and 
moderate scalability. Salcedo-Sanz et~al. introduced the Coral Reefs Optimisation algorithm~\cite{salcedo2014cro}, and 
later studies refined and applied CRO variants in broader optimisation contexts~\cite{emami2021cro, perezaracil2021cro}. CRO's reef-based competition mechanisms require moderate communication for spatial organisation and competitive interactions. The algorithm's balanced exploration--exploitation behaviour supports moderate energy efficiency, while its grid-based structure offers scalability dependent on grid size and reproduction~settings.

MPA demonstrates low communication dependency, high environmental adaptability, high energy efficiency, and 
good scalability. Faramarzi et~al. proposed the Marine Predators Algorithm~\cite{faramarzi2020mpa}, Al-Betar et~al. reported improved/extended variants~\cite{albetar2023mpa}, and 
Bujok provided a comparative evaluation across benchmark functions~\cite{bujok2023evaluation}. MPA's predator--prey dynamics enable effective coordination with minimal communication overhead, while its three-phase optimisation process adapts to different velocity ratios and environmental conditions. Its derivative-free formulation and limited parameter set contribute to energy efficiency, and 
its population-based structure supports~scaling.

This mapping reveals that no single algorithm dominates across all dimensions, highlighting the importance of mission-driven selection. Algorithms cluster into distinct regions of the classification space, with 
WOA and MPA occupying similar regions characterised by low communication dependency and high efficiency, while AFSA and CRO occupy positions with moderate communication demands and differing adaptability and scalability~characteristics.

\subsection{Comparative Evaluation of Marine-Specific Algorithms (AFSA, WOA, CRO, MPA)}

Systematic comparison of AFSA, WOA, CRO, and 
MPA reveals distinct performance characteristics that directly influence their suitability for underwater swarm robotics. A key limitation, however, is the absence of unified comparative studies that evaluate all four algorithms under consistent underwater-specific conditions, creating a gap in evidence-based algorithm selection~\cite{connor2021current}. Current insights rely on composite interpretations of separate studies that use different benchmarks, metrics, and 
environmental~assumptions.

In terms of communication dependency, WOA and MPA exhibit low communication requirements, relying primarily on global phase transitions or implicit coordination mechanisms rather than frequent inter-agent exchanges~\cite{mirjalili2016woa, faramarzi2020mpa}. In contrast, AFSA and CRO require moderate neighbourhood-level interactions to maintain swarm cohesion and support spatial competition~\cite{pourpanah2022afsa, salcedo2014cro}. This distinction is particularly critical in bandwidth-constrained acoustic communication~environments.

From the standpoint of energy and computational efficiency, all four algorithms maintain a per-iteration complexity of $O(N \cdot D)$, where $N$ is the number of agents and $D$ the dimensionality of the solution space. WOA and MPA benefit from minimal control parameters and simpler update mechanisms, while AFSA and CRO involve additional local decision rules and competition-based resource allocation that increase processing~overhead.

Regarding convergence and scalability, WOA's spiral bubble-net search enables fast convergence in smooth environments~\cite{mirjalili2016woa, nadimiShahraki2023woa}, while MPA's dual-phase Lévy and Brownian motion dynamics balance exploration and exploitation~\cite{faramarzi2020mpa}. AFSA maintains high population diversity through crowding factors but converges more gradually~\cite{pourpanah2022afsa}. CRO strikes a balance via competitive resource settlement among reef positions~\cite{salcedo2014cro}.

In terms of adaptability, AFSA and MPA demonstrate robust responses to environmental fluctuations due to their biologically grounded foraging dynamics~\cite{adeli2012review, mccluskey2021foraging}. WOA has been shown to adapt well to dynamic targets and current-driven changes~\cite{yan2022woa}, while CRO's ecosystem modelling supports moderate flexibility in multi-objective optimisation tasks~\cite{garciahernandez2020cro}.

Underwater application studies show differentiated use. AFSA has been applied to environmental monitoring, cooperative data collection, and 
Underwater Wireless Sensor Network routing~\cite{cai2023cooperative, kaur2024energy}. WOA variants have been employed in AUV path planning and current-aware navigation~\cite{yan2022woa}. MPA has been used for sensor network deployment and coverage optimisation, with 
emerging work in AUV control~\cite{albetar2023mpa, bujok2023evaluation}. CRO demonstrates strong performance in general optimisation but has limited underwater-specific validation~\cite{salcedo2014cro}.

Comparative evaluations across recent benchmarks have reinforced the distinct strengths of each algorithm. Pourpanah et~al.~\cite{pourpanah2022afsa} reported that AFSA maintains swarm diversity and accurate local search when extended through hybrid variants incorporating task-specific heuristics. Nadimi-Shahraki et~al.~\cite{nadimiShahraki2023woa} showed that WOA achieves fast convergence with relatively simple parameter settings, outperforming several traditional optimisers on a range of benchmark optimisation tasks. Salcedo-Sanz et~al.~\cite{salcedo2014cro} demonstrated that CRO provides competitive performance in multi-objective optimisation, reliably identifying near-optimal trade-offs in constrained search spaces. Bujok~\cite{bujok2023evaluation} finds that MPA, through its phase-based movement strategies, attains highly competitive absolute ranking scores across diverse benchmark functions. This indicates robust global search capability and~adaptability.

No single coordination model dominates across all performance metrics. Algorithm selection must reflect mission-specific objectives, communication limitations, and 
energy budgets. In this context, hybrid approaches that combine complementary algorithmic strengths represent a promising direction. For example, such approaches might integrate AFSA's local adaptability, WOA's global search efficiency, and 
MPA's phase transitions. These integrated frameworks can offer enhanced convergence, scalability, and 
robustness under uncertain underwater conditions. Empirical trends support integrating algorithmic features, for 
example combining MPA's Lévy-Brownian phases with AFSA's crowding mechanism, to 
build hybrid coordination models capable of addressing the multi-faceted challenges present in underwater swarm~robotics.

A consistent evaluation framework using shared metrics, mission profiles, and 
environmental models is needed to enable fair and transparent comparison, complementing the narrative synthesis of algorithm-specific strengths discussed in this~section.

\subsection{Integration of Algorithmic, Communication, and Hardware~Perspectives}

\textls[-5]{Effective underwater swarm robotics requires coherent integration of algorithms, communication strategies, and 
hardware capabilities. These domains are strongly interdependent and cannot be optimised independently. The classification framework provides a basis for analysing these interactions and aligning design choices with mission~requirements.}

Algorithm selection influences both communication and hardware requirements. Low-communication algorithms such as WOA and MPA reduce acoustic traffic and transmission energy, enabling simpler modem configurations and longer mission durations~\cite{mirjalili2016woa, faramarzi2020mpa}. Algorithms with moderate communication needs, such as AFSA and CRO, may require more capable modems or hybrid acoustic--optical architectures to maintain \mbox{performance~\cite{pourpanah2022afsa, salcedo2014cro}}.

Communication constraints shape algorithm feasibility and swarm architecture. Severe bandwidth limits favour algorithms that rely on local sensing and minimal message exchange; intermittent connectivity requires fault-tolerant coordination that remains functional during communication blackouts~\cite{connor2021current, gussen2021optimisation}. Hardware platforms must integrate suitable communication modalities and antenna designs to support these strategies, influencing size, cost, and 
payload~allocation.

Hardware capabilities, in 
turn, constrain algorithm and communication choices. Compact vehicles with limited payload and power cannot support heavy sensing loads or multiple high-power modems, favouring algorithms that operate effectively with sparse sensing and low-rate communication~\cite{zhang2023autonomous, osterloh2012monsun}. Energy limitations demand algorithms and protocols that minimise communication and computation, reinforcing the importance of energy-aware design~\cite{guo2019sustainable, khan2025hybrid}.

These interdependencies illustrate mission-dependent trade-offs. Long-duration environmental monitoring favours low-communication, energy-efficient algorithms on modest hardware. High-precision inspection tasks may prioritise formation fidelity and sensing accuracy, accepting higher communication and energy costs. Complex multi-phase missions may require adaptive systems capable of reconfiguring algorithm parameters, communication modes, and 
sensing policies in response to changing objectives and~conditions.

Cross-layer optimisation approaches offer a way to manage these trade-offs holistically~\cite{gussen2021optimisation, khan2025hybrid}. Adaptive physical-layer methods that tune modulation and power settings based on channel conditions exemplify coordination between communication and energy objectives. Energy-aware trajectory planning that leverages currents illustrates how hardware and environment can be exploited to satisfy algorithmic goals while minimising energy use. Such integrated designs represent a key step towards practical, large-scale underwater swarm~deployment.

\subsection{Summary of Key Trends and~Limitations}

Synthesising findings across bio-inspired coordination, communication, and 
system design reveals several key trends in underwater swarm robotics, along with persistent limitations that constrain current~progress.

First, there is a clear trend towards marine-specific bio-inspired algorithms that align biological mechanisms with underwater constraints. AFSA, WOA, CRO, and 
MPA reflect increasing recognition that biologically grounded strategies can address communication limits, environmental variability, and 
energy constraints more naturally than generic optimisation~methods.

Second, communication research has progressed from simple acoustic links to adaptive and hybrid architectures that integrate OFDM, multimodal links, and 
bio-inspired signalling. Field deployments have demonstrated the feasibility of adaptive physical-layer optimisation and TDMA-based scheduling in real underwater environments~\cite{gussen2021optimisation}.

Third, hardware trends point towards modular, cost-effective platforms capable of supporting swarm-scale deployment. Diverse design strategies ranging from high-capability platforms to compact systems illustrate different approaches tuned to various mission profiles~\cite{gussen2021optimisation}.

A critical limitation is the scarcity of real-world data and longitudinal field trials. The literature is dominated by simulation studies, with 
limited progression to empirical validation in representative environments~\cite{connor2021current}. This restricts understanding of how algorithms, communication strategies, and 
hardware behave under real ocean conditions, where uncertainty and unmodelled effects are~significant.

Future research directions highlighted by this synthesis include the following:  
(i) empirical validation of swarm systems in realistic marine environments;  
(ii) development of hybrid algorithms that combine complementary strengths from AFSA, WOA, CRO, and 
MPA;  
(iii)~advanced communication protocols tailored to multi-vehicle autonomy;  
(iv) standardised benchmarking and reporting frameworks; and  
(v) integrated design methodologies that explicitly account for cross-layer~interdependencies.  

The classification framework presented here provides a structured foundation for addressing these challenges by mapping algorithms across communication dependency, environmental adaptability, energy efficiency, and 
swarm scalability. This mapping supports systematic, mission-driven decision-making and helps to bridge current fragmentation in underwater swarm robotics research, guiding progress towards robust, scalable, and 
operational marine swarm~systems.

\section{Discussion and Research~Outlook}
\label{discussion}

This review has synthesised findings across bio-inspired coordination mechanisms, communication strategies, and 
system design considerations, revealing both substantial progress and persistent challenges in underwater swarm robotics. While individual domains have advanced, the 
integration of these components into cohesive operational systems remains limited. The discussion covers cross-disciplinary insights emerging from the convergence of biological, engineering, and 
computational research. It also identifies technological gaps constraining practical deployment, outlines future research directions, and 
considers pathways towards intelligent and sustainable marine operations enabled by underwater swarm~robotics.

\subsection{Cross-Disciplinary Insights and Converging~Trends}

Synthesis of findings across biology, control theory, marine engineering, and 
computer science reveals several converging trends that characterise the evolution of underwater swarm robotics. These cross-disciplinary insights highlight how biological inspiration, communication constraints, and 
hardware capabilities interact to shape system design and~performance.

A major trend is the increasing recognition that marine-specific bio-inspired algorithms provide natural compatibility with underwater operational constraints. The development of AFSA, WOA, CRO, and 
MPA demonstrates how biological mechanisms evolved in marine environments align with robotic requirements for communication-limited coordination, environmental adaptability, and 
energy efficiency~\cite{zhao2025bioinspired}. This biological--engineering convergence reflects the growing understanding that natural systems have already solved many of the challenges confronting robotic swarms, offering validated design principles transferable to technological~systems.

Communication research shows convergence towards adaptive and hybrid approaches that address bandwidth limitations through intelligent resource management. The emergence of adaptive physical-layer methodologies, hybrid acoustic-optical architectures, and 
bio-inspired communication protocols demonstrates recognition that fixed communication strategies cannot accommodate the variability of underwater environments~\cite{gussen2021optimisation, lodovisi2018performance}. These adaptive approaches parallel biological systems that dynamically adjust behaviour in response to environmental cues, creating synergy between bio-inspired algorithms and communication~strategies.

Hardware design trends demonstrate convergence towards modular, cost-effective platforms that support scalable swarm deployment while maintaining operational capability. The contrast between comprehensive platforms and lightweight solutions illustrates how distinct design philosophies target different mission requirements~\cite{gussen2021optimisation}. This diversity reflects the realisation that swarm scalability depends on optimising trade-offs between individual vehicle capability, endurance, and 
manufacturing~cost.

A critical converging insight is the interdependence between algorithmic, communication, and 
hardware domains. Algorithms cannot be chosen independently of communication constraints; communication protocols must accommodate algorithmic requirements; and hardware must support both computation and communication functions. This interdependence produces complex design spaces requiring integrated optimisation rather than isolated component design. Such integration underscores the importance of interdisciplinary collaboration and systematic frameworks addressing these~relationships.

\subsection{Technological Gaps and Practical~Challenges}

Despite notable progress, several technological gaps and practical challenges constrain the transition of underwater swarm robotics from research to operational~deployment.

A fundamental gap is the scarcity of empirical validation studies in realistic underwater conditions. Most published work remains simulation-based, with 
limited progression to controlled or field trials~\cite{connor2021current}. This simulation-to-field gap reduces confidence in performance claims and restricts understanding of system behaviour under true ocean variability. In real ocean conditions, uncertainty, unmodelled dynamics, and 
environmental heterogeneity strongly influence outcomes. Field testing is constrained by cost, logistical complexity, and 
safety concerns, forming a persistent barrier to empirical~progress.

The absence of unified comparative evaluation across marine-specific algorithms represents another major gap. Current comparisons rely on heterogeneous metrics and test conditions, making direct performance benchmarking unreliable~\cite{connor2021current}. Practitioners therefore lack standardised references for algorithm selection, leading to reliance on case-specific studies that may not generalise across missions or~environments.

It is also important to acknowledge an ongoing methodological controversy in bio-inspired optimisation and swarm literature: new algorithm names are frequently introduced even when the underlying methodology is closely related to existing metaheuristics or represents a relatively minor modification (e.g., re-parameterisation, operator substitution, or 
problem-specific hybridisation). In this review, we treat algorithm names primarily as a taxonomy for communicating biological inspiration and design intent, rather than as a claim of fundamental novelty. Where possible, we emphasise substantive originality in terms of (i) clearly defined update operators or interaction rules; (ii) explicit modelling of underwater constraints (communication delay, localisation uncertainty, energy limits, hydrodynamics); and (iii) transparent empirical evidence (comparisons on shared benchmarks, ablations, and 
reporting of sensitivity to hyperparameters). This perspective motivates the need for standardised benchmarking and reporting so that the contribution of newly named algorithms can be assessed on measurable performance and reproducibility rather than~nomenclature.

Integration challenges remain a central practical limitation. While algorithms, communication protocols, and 
hardware platforms each perform effectively in isolation, their integration into operational systems introduces complex interdependencies~\cite{wubben2020challenges, gussen2021optimisation}. The absence of comprehensive system-level studies limits understanding of emergent behaviours, coordination robustness, and 
collective fault tolerance. These factors are critical to real-world reliability. Understanding emergent properties and failure modes in fully integrated swarms is essential for operational~reliability.

Research into hybrid algorithmic frameworks combining complementary strengths from existing approaches remains underdeveloped. AFSA-WOA and CRO-MPA hybridisation represents a promising avenue for addressing multi-objective missions that exceed single-algorithm capabilities~\cite{khan2025hybrid, ismail2021systematic}, yet systematic frameworks for designing and validating such hybrids are~lacking.

Another core limitation arises from the temporal simplicity of current algorithms. AFSA models schooling and feeding, WOA represents bubble-net hunting, CRO captures spawning dynamics, and 
MPA encodes predator--prey pursuit. However, real marine organisms display multiple behavioural states that shift in response to stimuli, seasonality, and 
ecological context~\cite{clapham2000humpback, mccluskey2021foraging}. Current algorithms capture only static behaviours, limiting adaptability to dynamic mission phases requiring behaviour~transitions.

From a literature coverage perspective, this review is necessarily selective. Its focus on marine-specific bio-inspired algorithms such as AFSA, WOA, CRO, and 
MPA means that many general-purpose bio-inspired and swarm optimisation methods originally developed for terrestrial and aerial domains are only briefly mentioned or fall outside the primary scope. This includes Particle Swarm Optimisation (PSO) and Artificial Bee Colony (ABC) algorithms that have been adapted to underwater applications~\cite{guo2009pso, prasath2022abc}. Likewise, other marine-inspired optimisation frameworks such as Chaotic Krill Herd algorithms, Fish School Search, and 
Dolphin Echolocation-based optimisation~\cite{wang2014ckh, bastos2008fss, kaveh2013dea} are not examined in detail here. These additional algorithm families represent important avenues for future comparative studies and hybrid designs that combine marine-specific and more general bio-inspired optimisation~strategies.

A critical gap is the lack of standardisation, which constrains interoperability, reproducibility, and 
comparative evaluation. Divergent hardware interfaces, communication stacks, and 
software frameworks hinder modularity and prevent consistent cross-platform evaluation~\cite{gussen2021optimisation}. The absence of standardised benchmarking protocols, unified evaluation metrics, and 
shared test conditions makes it difficult to compare algorithms and system architectures fairly. Standardised testing environments, data formats, interface protocols, and 
evaluation frameworks are essential for enabling system interoperability and facilitating algorithm selection. Such standardisation would accelerate cumulative progress towards evidence-based design~\cite{connor2021current, milner2023swarm}.

\subsection{Future Research Directions in Bio-Inspired Marine~Robotics}

Addressing these limitations requires coordinated, multi-domain research efforts targeting algorithmic innovation, communication efficiency, hardware integration, and 
validation~methodology.

In terms of algorithm development and optimisation, empirical validation of marine-specific algorithms in real underwater environments remains a critical priority. Hybrid algorithms that combine complementary mechanisms from AFSA, WOA, CRO, and 
MPA could meet complex, multi-objective mission requirements~\cite{khan2025hybrid, ismail2021systematic}. Adaptive parameter tuning and context-aware switching between behavioural modes could enable algorithms to dynamically respond to environmental and mission changes. Incorporating behavioural transitions analogous to biological state shifts could substantially increase adaptability and resilience. Such transitions might include exploration, aggregation, or 
avoidance modes that shift in response to mission~requirements.

Future work in communication and networking must develop next-generation communication protocols that address the combined challenges of limited bandwidth, high latency, and 
variable reliability~\cite{wubben2020challenges}. Adaptive, energy-efficient strategies capable of real-time adjustment to environmental conditions are required. Integration of multiple modalities (acoustic, optical, RF, Magnetic Induction) offers a pathway to robust hybrid networks. Bio-inspired communication mechanisms that exploit environmental cues or implicit coordination could reduce data traffic while maintaining swarm~coherence.

From a hardware and system design perspective, autonomous and modular hardware architectures are needed to support scalable swarm deployment. Compact, reconfigurable platforms with interchangeable payloads could enable cost-effective adaptation across missions. Advanced sensor integration and distributed perception systems would improve situational awareness. Energy harvesting, battery innovation, and 
energy-aware propulsion will underpin longer missions, while cost-performance optimisation can enable deployment of large swarms for wide-area~monitoring.

\textls[-11]{Regarding evaluation and validation, real-time monitoring systems providing continuous performance metrics can enhance adaptive control and safety. Progressive validation (starting with simulation, advancing through Hardware-in-the-Loop, and 
culminating in controlled field trials) offers a structured pathway to operational readiness. Transparent reporting standards and open datasets would further strengthen reproducibility and~collaboration.}

For integration and system-level design, integrated methodologies that jointly optimise algorithmic, communication, and 
hardware layers will be essential for practical deployment~\cite{gussen2021optimisation}. Cross-layer optimisation can improve energy use and coordination fidelity, while standardised interfaces will facilitate modular development and technology transfer. Large-scale system studies addressing emergent behaviours, cooperative fault handling, and 
resilience under realistic conditions remain a pressing research~frontier.

In terms of applications and commercialisation, translating swarm robotics into commercial practice will depend on validated applications such as infrastructure inspection, resource monitoring, and 
search-and-rescue operations. Economic feasibility studies, environmental impact assessments, and 
regulatory frameworks are needed to ensure sustainable and responsible deployment. Collaboration with marine industries could align technological development with real operational~needs.

\subsection{Limitations}

While this review aimed to integrate bio-inspired coordination, communication, and 
system design perspectives, several limitations should be acknowledged. The literature considered was limited to English-language publications from selected databases, which may exclude relevant studies in other languages or formats. No formal quantitative quality appraisal was performed; instead, studies were assessed qualitatively based on relevance and methodological clarity. Finally, most of the reviewed works rely on simulation-based validation with limited real-world or long-duration field trials, which constrains the generalisability of findings to complex marine environments. These factors highlight the need for standardised benchmarking, empirical evaluation, and 
open datasets in future underwater swarm robotics research. In addition, some topical areas may have been under-represented due to venue and format choices (e.g., conference-centric underwater robotics and networking work, non-English regional literature, and 
applied industry/technical~reports).

\subsection{Towards Intelligent and Sustainable Marine~Operations}

Advancing underwater swarm robotics towards intelligent and sustainable operation requires both conceptual and technological evolution. Intelligence refers to systems capable of autonomous adaptation, dynamic reconfiguration, and 
robust decision-making under uncertainty. Sustainability encompasses energy efficiency, environmental responsibility, and 
long-term operational~viability.

Future intelligent swarm systems will employ multi-modal algorithms incorporating behavioural transitions that enable dynamic adaptation to mission objectives, environmental conditions, and 
emergent events. Such systems could autonomously switch between exploration, inspection, or 
emergency modes in response to sensor feedback. Integration of adaptive learning and parameter tuning will allow continual self-optimisation based on operational experience. This capability would enable systems to improve performance over time through accumulated~experience.

Sustainability depends on energy-aware design across all layers of the system. Bio-inspired algorithms that minimise communication, adaptive protocols that optimise energy expenditure, and 
current-assisted trajectory planning collectively extend mission endurance~\cite{zhang2024energy, cai2021stream}. Energy harvesting, efficient power management, and 
advanced storage technologies will further support long-duration operations required for persistent ocean observation. Applications such as offshore wind farm monitoring, sub-sea pipeline inspection, and 
marine disaster response demonstrate the transformative potential of intelligent swarm systems for sustainable ocean operations. These applications highlight how energy-efficient swarm systems can enable persistent, distributed monitoring at scales previously~impractical.

Integration of swarm robotics into broader marine science and engineering objectives promises transformative benefits. Persistent, distributed observation networks could enhance climate monitoring, biodiversity assessment, and 
pollution detection. Multi-vehicle inspection and maintenance operations could improve infrastructure reliability while reducing cost and environmental~footprint.

\textls[-11]{Realising these capabilities requires sustained interdisciplinary collaboration linking marine engineering, computer science, and 
environmental science~\cite{zhao2025bioinspired}. Education and knowledge-transfer} frameworks are equally vital to develop expertise and foster collaboration between academia, industry, and 
policy bodies. Open-source platforms and community-driven datasets could further accelerate progress and promote equitable participation in marine robotics~research.

The pathway towards intelligent and sustainable marine operations demands incremental, evidence-based progress across all these domains. The classification framework and synthesis presented in this review provide structured foundations for addressing the field's fragmentation. This framework enables systematic evaluation and selection of coordination strategies that balance competing objectives. By integrating algorithmic, communication, and 
hardware considerations, underwater swarm systems can evolve towards robust, scalable, and 
environmentally sustainable operations. Such evolution would advance both technological capability and marine~stewardship.

\section{Conclusions}

This review synthesises research on bio-inspired coordination, communication, and 
system design for underwater swarm robotics, drawing on 446 peer-reviewed articles from 2001 to 2025, with 
the majority of publications concentrated within the past decade reflecting the field's recent rapid growth. The principal contribution is a four-dimensional classification framework evaluating algorithms across communication dependency, environmental adaptability, energy efficiency, and 
swarm scalability, providing a systematic basis for mission-driven algorithm~selection.

Comparative analysis of AFSA, WOA, CRO, and 
MPA reveals distinct trade-offs: WOA and MPA exhibit low communication dependency and high energy efficiency, making them suited to bandwidth-constrained acoustic environments, while AFSA offers strong environmental adaptability through local neighbourhood interactions. The review identifies that adaptive physical-layer communication strategies can reduce transmission energy by 30--50\% and 
that energy-aware path planning can lower propulsion costs by 20--30\%. Field-validated platforms such as COMET and NemoSens demonstrate that decentralised TDMA-based coordination is achievable in real underwater~conditions.

Three critical gaps constrain progress: (i) the dominance of simulation-based validation with limited field trials, (ii) the absence of standardised benchmarks for cross-algorithm comparison, and 
(iii) insufficient integration of algorithmic, communication, and 
hardware layers into cohesive systems. Addressing these gaps requires hybrid algorithm development combining complementary strengths, empirical validation under realistic ocean conditions, and 
standardised evaluation frameworks. The classification framework presented here provides a foundation for such progress, guiding the evolution of underwater swarm systems towards robust, scalable, and 
energy-efficient marine~operations.

\vspace{6pt}

\authorcontributions{Conceptualisation, S.R., A.S. and S.M.; methodology, S.R.; formal analysis, S.R.; investigation, S.R.; resources, S.R. and S.M.; data curation, S.R.; writing---original draft preparation, S.R.; writing---review and editing, S.M. and A.S.; supervision, S.M. and A.S. All authors have read and agreed to the published version of the manuscript.}

\funding{This research received no external~funding.}

%

\dataavailability{No new data were created or analysed in this study. Data sharing is not applicable to this article.} 

\acknowledgments{The authors would like to acknowledge \hl{Eric} 
 Pardede, Alex Tomy, Kiki Adhinugraha, Tony de Souza-Daw, Fernando Galetto, and 
Adam Console, each of whom has contributed uniquely to the development of this research article. Their invaluable advice, encouragement, and 
constructive feedback have enriched our thinking and broadened our perspective. Whether through technical discussions or strategic guidance, their collective input has shaped the outcomes of this~\hl{work}.}   

\conflictsofinterest{The authors declare no conflicts of~interest.}

\abbreviations{Abbreviations}{
The following abbreviations are used in this manuscript:\\

\noindent 
\begin{tabular}{@{}m{1.8cm}<{\raggedright}m{10.5cm}<{\raggedright}}
AFSA & Artificial Fish Swarm Algorithm\\
ABC & Artificial Bee Colony\\
AUV & Autonomous Underwater Vehicle\\
BER & Bit-Error Rate\\
CKH & Chaotic Krill Herd\\
CRO & Coral Reef Optimisation\\
DEA & Dolphin Echolocation-based Optimisation\\
DTN & Delay-Tolerant Network\\
DVL & Doppler Velocity Log\\
FAD & Fish Aggregating Devices\\
FSS & Fish School Search\\
GPS & Global Positioning System\\
HIL & Hardware-in-the-Loop\\
IMU & Inertial Measurement Unit\\
INS & Inertial Navigation System\\
\end{tabular}

\noindent 
\begin{tabular}{@{}m{1.8cm}<{\raggedright}m{10.5cm}<{\raggedright}}
LBL & Long Baseline\\
MASF & Multi-Agent System Frameworks\\
MI & Magnetic Induction\\
MPA & Marine Predators Algorithm\\
MRS & Marine Robotic Systems\\
OFDM & Orthogonal Frequency-Division Multiplexing\\
PDR & Packet Delivery Ratio\\
PSO & Particle Swarm Optimisation\\
RF & Radio Frequency\\
ROS & Robot Operating System\\
ROV & Remotely Operated Vehicle\\
RSF & Robotics Software Frameworks\\
TDMA & Time-Division Multiple Access\\
USBL & Ultra-Short Baseline\\
USV & Unmanned Surface Vessel\\
UUV & Unmanned Underwater Vehicle\\
UWSN & Underwater Wireless Sensor Network\\
WOA & Whale Optimisation Algorithm
\end{tabular}
}

\begin{adjustwidth}{-\extralength}{0cm}

\reftitle{References}

\PublishersNote{}
\end{adjustwidth}

\begin{thebibliography}{999}
\bibitem[{Di Ciaccio} and Troisi(2021)]{di_ciaccio2021monitoring}
{Di Ciaccio}, F.; Troisi, S.
Monitoring marine environments with Autonomous Underwater Vehicles: A bibliometric analysis.
{\em Results Eng.} {\bf 2021}, {\em 9},~100205. [\href{http://doi.org/10.1016/j.rineng.2021.100205}{CrossRef}]
\bibitem[Franke et~al.(2020)Franke, Blenckner, Duarte, Ott, Fleming, Antia, Reusch, Bertram, Hein, Kronfeld-Goharani, Dierking, Kuhn, Sato, {van Doorn}, Wall, Schartau, Karez, Crowder, Keller, Engel, Hentschel, and Prigge]{franke2020operationalizing}
Franke, A.; Blenckner, T.; Duarte, C.M.; Ott, K.; Fleming, L.E.; Antia, A.; Reusch, T.B.; Bertram, C.; Hein, J.; Kronfeld-Goharani, U.; et~al.
Operationalizing Ocean Health: Toward Integrated Research on Ocean Health and Recovery to Achieve Ocean Sustainability.
{\em ONE Earth} {\bf 2020}, {\em 2},~557--565. [\href{http://dx.doi.org/10.1016/j.oneear.2020.05.013}{CrossRef}]
\bibitem[Petillot et~al.(2019)Petillot, Antonelli, Casalino, and Ferreira]{petillot2019underwater}
Petillot, Y.R.; Antonelli, G.; Casalino, G.; Ferreira, F.
Underwater Robots: From Remotely Operated Vehicles to Intervention-Autonomous Underwater Vehicles.
{\em IEEE Robot. Autom. Mag.} {\bf 2019}, {\em 26},~94--101. [\href{http://dx.doi.org/10.1109/MRA.2019.2908063}{CrossRef}]
\bibitem[Zhao et~al.(2025)Zhao, Yang, Tang, Yang, Dong, Xi, Zou, Xu, Li, and Wang]{zhao2025bioinspired}
Zhao, Q.; Yang, T.; Tang, G.; Yang, Y.; Dong, F.; Xi, Z.; Zou, Y.; Xu, M.; Li, S.; Wang, C.
Bio-inspired swarm of underwater robots: A review.
{\em Bioinspir. Biomim.} {\bf 2025}, {\em 20},~041002. [\href{http://dx.doi.org/10.1088/1748-3190/ade215}{CrossRef}]
\bibitem[Brambilla et~al.(2013)Brambilla, Ferrante, Birattari, and Dorigo]{brambilla2013swarm}
Brambilla, M.; Ferrante, E.; Birattari, M.; Dorigo, M.
Swarm robotics: A review from the swarm engineering perspective.
{\em Swarm Intell.} {\bf 2013}, {\em 7},~1--41. [\href{http://dx.doi.org/10.1007/s11721-012-0075-2}{CrossRef}]
\bibitem[Ismail and Hamami(2021)]{ismail2021systematic}
Ismail, Z.H.; Hamami, M.G.M.
Systematic Literature Review of Swarm Robotics Strategies Applied to Target Search Problem with Environment Constraints.
{\em Appl. Sci.} {\bf 2021}, {\em 11},~2383. [\href{http://dx.doi.org/10.3390/app11052383}{CrossRef}]
\bibitem[Connor et~al.(2021)Connor, Champion, and Joordens]{connor2021current}
Connor, J.; Champion, B.; Joordens, M.A.
Current Algorithms, Communication Methods and Designs for Underwater Swarm Robotics: A Review.
{\em IEEE Sens. J.} {\bf 2021}, {\em 21},~153--169. [\href{http://dx.doi.org/10.1109/JSEN.2020.3013265}{CrossRef}]
\bibitem[Chen et~al.(2023)Chen, Wang, Liu, Yu, Yue, Song, and Lin]{chen2023modelling}
Chen, P.; Wang, F.; Liu, S.; Yu, Y.; Yue, S.; Song, Y.; Lin, Y.
Modeling collective behavior for fish school with deep Q-networks.
{\em IEEE Access} {\bf 2023}, {\em 11},~36630--36641. [\href{http://dx.doi.org/10.1109/ACCESS.2023.3263237}{CrossRef}]
\bibitem[Dorigo et~al.(2021)Dorigo, Theraulaz, and Trianni]{dorigo2021swarm}
Dorigo, M.; Theraulaz, G.; Trianni, V.
Swarm Robotics: Past, Present, and Future [Point of View].
{\em Proc. IEEE} {\bf 2021}, {\em 109},~1152--1165. [\href{http://dx.doi.org/10.1109/JPROC.2021.3072740}{CrossRef}]
\bibitem[Arksey and O'Malley(2005)]{arksey2005scoping}
Arksey, H.; O'Malley, L.
Scoping studies: Towards a methodological framework.
{\em Int. J. Soc. Res. Methodol.} {\bf 2005}, {\em 8},~19--32. [\href{http://dx.doi.org/10.1080/1364557032000119616}{CrossRef}]
\bibitem[Braun and Clarke(2006)]{braun2006thematic}
Braun, V.; Clarke, V.
Using thematic analysis in psychology.
{\em Qual. Res. Psychol.} {\bf 2006}, {\em 3},~77--101. [\href{http://dx.doi.org/10.1191/1478088706qp063oa}{CrossRef}]
\bibitem[Jabareen(2009)]{jabareen2009building}
Jabareen, Y.
Building a conceptual framework: Philosophy, definitions, and procedure.
{\em Int. J. Qual. Methods} {\bf 2009}, {\em 8},~49--62. [\href{http://dx.doi.org/10.1177/160940690900800406}{CrossRef}]
\bibitem[{Fogh Sørensen} et~al.(2025){Fogh Sørensen}, Mai, {von Benzon}, Liniger, and Pedersen]{sorensen2025localization}
{Fogh Sørensen}, F.; Mai, C.; {von Benzon}, M.; Liniger, J.; Pedersen, S.
The localization problem for underwater vehicles: An overview of operational solutions.
{\em Ocean Eng.} {\bf 2025}, {\em 330},~121173. [\href{http://dx.doi.org/10.1016/j.oceaneng.2025.121173}{CrossRef}]
\bibitem[Potokar et~al.(2021)Potokar, Norman, and Mangelson]{potokar2021invariant}
Potokar, E.R.; Norman, K.; Mangelson, J.G.
Invariant Extended Kalman Filtering for Underwater Navigation.
{\em IEEE Robot. Autom. Lett.} {\bf 2021}, {\em 6},~5792--5799. [\href{http://dx.doi.org/10.1109/LRA.2021.3085167}{CrossRef}]
\bibitem[Ligęza(2023)]{ligeza2023reconstructing}
Ligęza, P.
Reconstructing the trajectory of the object's motion on the basis of measuring the components of its velocity.
{\em Measurement} {\bf 2023}, {\em 221},~113546. [\href{http://dx.doi.org/10.1016/j.measurement.2023.113546}{CrossRef}]
\bibitem[Awan et~al.(2019)Awan, Shah, Iqbal, Gillani, Ahmad, and Nam]{awan2019underwater}
Awan, K.M.; Shah, P.A.; Iqbal, K.; Gillani, S.; Ahmad, W.; Nam, Y.
Underwater Wireless Sensor Networks: A Review of Recent Issues and Challenges.
{\em Wirel. Commun. Mob. Comput.} {\bf 2019}, {\em 2019},~6470359. [\href{http://dx.doi.org/10.1155/2019/6470359}{CrossRef}]
\bibitem[Saeed et~al.(2019)Saeed, Celik, Al-Naffouri, and Alouini]{saeed2019underwater}
Saeed, N.; Celik, A.; Al-Naffouri, T.Y.; Alouini, M.S.
Underwater optical wireless communications, networking, and localization: A survey.
{\em Ad Hoc Netw.} {\bf 2019}, {\em 94},~101935. [\href{http://dx.doi.org/10.1016/j.adhoc.2019.101935}{CrossRef}]
\bibitem[Alahmad et~al.(2023)Alahmad, Ishii, Nishida, Fukumoto, and Matsushima]{alahmad2023experimental}
Alahmad, R.; Ishii, K.; Nishida, Y.; Fukumoto, Y.; Matsushima, T.
Experimental Study of Underwater RF Communication for Live Video Transmission for AUVs Application.
{\em J. Robot. Netw. Artif. Life} {\bf 2023}, {\em 10},~84--90. [\href{http://dx.doi.org/10.57417/jrnal.10.1_84}{CrossRef}]
\bibitem[Zhang and Lauder(2024)]{zhang2024energy}
Zhang, Y.; Lauder, G.V.
Energy conservation by group dynamics in schooling fish.
{\em eLife} {\bf 2024}, {\em 12},~RP90352. [\href{http://dx.doi.org/10.7554/eLife.90352}{CrossRef}]
\bibitem[Wu et~al.(2025)Wu, Deng, and et~al.]{wu2025minimalistic}
Wu, X.; Deng, X.; Wen, B.; Yue, S.; Shao, S.; Zhang, F.; Wang, F.; Lin, Y.
A Minimalistic and Decentralised Approach to Formation Control for Crowded UUV Swarms Inspired by Fish Schooling.
{\em J. Bionic Eng.} {\bf 2025}, \emph{22}, 2646--2659. [\href{http://dx.doi.org/10.1007/s42235-025-00766-w}{CrossRef}]
\bibitem[Geng et~al.(2019)Geng, Li, and Xu]{geng2019pheromone}
Geng, C.; Li, G.; Xu, H.
Pheromone Inspired with Directional Variable on Underwater Robot Swarm.
In Proceedings of the 2019 Chinese Control And Decision Conference (CCDC), \hl{Nanchang, China, 3--5 June} 
2019; pp.~5036--5040. [\href{http://dx.doi.org/10.1109/CCDC.2019.8833464}{CrossRef}]
\bibitem[Li et~al.(2019)Li, Chen, and et~al.]{li2019pheromone}
Li, G.; Chen, C.; Geng, C.; Li, M.; Xu, H.; Lin, Y.
A pheromone-inspired monitoring strategy using a swarm of underwater robots.
{\em Sensors} {\bf 2019}, {\em 19},~4089. [\href{http://dx.doi.org/10.3390/s19194089}{CrossRef}]
\bibitem[Cai et~al.(2023)Cai, Liu, Zhang, and Wang]{cai2023cooperative}
Cai, W.; Liu, Z.; Zhang, M.; Wang, C.
Cooperative Artificial Intelligence for underwater robotic swarm.
{\em Robot. Auton. Syst.} {\bf 2023}, {\em 164},~104410. [\href{http://dx.doi.org/10.1016/j.robot.2023.104410}{CrossRef}]
\bibitem[Yan et~al.(2022)Yan, Zhang, Zeng, and Tang]{yan2022woa}
Yan, Z.; Zhang, J.; Zeng, J.; Tang, J.
Three-dimensional path planning for autonomous underwater vehicles based on a whale optimization algorithm.
{\em Ocean Eng.} {\bf 2022}, {\em 250},~111070. [\href{http://dx.doi.org/10.1016/j.oceaneng.2022.111070}{CrossRef}]
\bibitem[Emami et~al.(2021)Emami, Nazif, Mousavi, Karami, and Daccache]{emami2021cro}
Emami, M.; Nazif, S.; Mousavi, S.F.; Karami, H.; Daccache, A.
A hybrid constrained coral reefs optimization algorithm with machine learning for optimizing multi-reservoir systems operation.
{\em J. Environ. Manag.} {\bf 2021}, {\em 286},~112250. [\href{http://dx.doi.org/10.1016/j.jenvman.2021.112250}{CrossRef}]
\bibitem[Ramesh et~al.(2025)Ramesh, Mann, and Stumpf]{ramesh2025naupliusoptimisationautonomoushydrodynamics}
Ramesh, S.; Mann, S.; Stumpf, A.
Nauplius Optimisation for Autonomous Hydrodynamics.
\emph{arXiv} \textbf{2025}. [\href{http://dx.doi.org/10.48550/arXiv.2510.15350}{CrossRef}]
\bibitem[Darvishpoor et~al.(2023)Darvishpoor, Darvishpour, Escarcega, and Hassanalian]{darvishpoor2023nature}
Darvishpoor, S.; Darvishpour, A.; Escarcega, M.; Hassanalian, M.
Nature-Inspired Algorithms from Oceans to Space: A Comprehensive Review of Heuristic and Meta-Heuristic Optimization Algorithms and Their Potential Applications in Drones.
{\em Drones} {\bf 2023}, {\em 7},~427. [\href{http://dx.doi.org/10.3390/drones7070427}{CrossRef}]
\bibitem[Ni et~al.(2018)Ni, Yang, Wu, and Fan]{ni2018improved}
Ni, J.; Yang, L.; Wu, L.; Fan, X.
An Improved Spinal Neural System-Based Approach for Heterogeneous AUVs Cooperative Hunting.
{\em Int. J. Fuzzy Syst.} {\bf 2018}, {\em 20},~672--686. [\href{http://dx.doi.org/10.1007/s40815-017-0395-x}{CrossRef}]
\bibitem[Guerrero-Criollo et~al.(2023)Guerrero-Criollo, Casta\~{n}o L\'{o}pez, Hurtado-L\'{o}pez, and Ramirez-Moreno]{guerrerocriollo2023bioinspired}
Guerrero-Criollo, R.J.; Casta\~{n}o L\'{o}pez, J.A.; Hurtado-L\'{o}pez, J.; Ramirez-Moreno, D.F.
Bio-Inspired Neural Networks for Decision-Making Mechanisms and Neuromodulation for Motor Control in a Differential Robot.
{\em Front. Neurorobot.} {\bf 2023}, {\em 17},~1078074. [\href{http://dx.doi.org/10.3389/fnbot.2023.1078074}{CrossRef}]
\bibitem[Ni et~al.(2016)Ni, Li, Hua, and Yang]{ni2016bioinspired}
Ni, J.; Li, X.; Hua, M.; Yang, S.X.
Bioinspired Neural Network-Based Q-Learning Approach for Robot Path Planning in Unknown Environments.
{\em Int. J. Robot. Autom.} {\bf 2016}, {\em 31}, 464--474. [\href{http://dx.doi.org/10.2316/Journal.206.2016.6.206-4526}{CrossRef}]
\bibitem[Kaur et~al.(2024)Kaur, Kaur, Singh, et~al.]{kaur2024energy}
Kaur, P.; Kaur, K.; Singh, K.; Saleem, K.; Ur Rehman, A.; Gupta, R.; Hussen Adem, S.
Energy-efficient artificial fish swarm-based clustering protocol for enhancing network lifetime in underwater wireless sensor networks.
{\em J. Wirel. Commun. Netw.} {\bf 2024}, {\em 2024},~92. [\href{http://dx.doi.org/10.1186/s13638-024-02422-z}{CrossRef}]
\bibitem[Adeli(2012)]{adeli2012review}
Adeli, A.
A Review of Artificial Fish Swarm Optimization Methods and Applications.
{\em Int. J. Smart Sens. Intell. Syst.} {\bf 2012}, \emph{5}, 107--148. [\href{http://dx.doi.org/10.21307/ijssis-2017-474}{CrossRef}]
\bibitem[Pourpanah et~al.(2022)Pourpanah, Wang, Lim, Wang, and Yazdani]{pourpanah2022afsa}
Pourpanah, F.; Wang, R.; Lim, C.; Wang, X.Z.; Yazdani, D.
A review of artificial fish swarm algorithms: Recent advances and applications.
{\em Artif. Intell. Rev.} {\bf 2022}, {\em 56},~1867--1903. [\href{http://dx.doi.org/10.1007/s10462-022-10214-4}{CrossRef}]
\bibitem[Peraza et~al.(2022)Peraza, Ochoa, Amador, and Castillo]{peraza2022afsa}
Peraza, C.; Ochoa, P.; Amador, L.; Castillo, O.
Artificial Fish Swarm Algorithm for the Optimization of a Benchmark Set of Functions. In {\em New Perspectives on Hybrid Intelligent System Design Based on Fuzzy Logic, Neural Networks and Metaheuristics}; Castillo, O., Melin, P., Eds.; Springer International Publishing: Cham, Switzerland, 2022; pp.~77--92. [\href{http://dx.doi.org/10.1007/978-3-031-08266-5_6}{CrossRef}]
\bibitem[Mirjalili and Lewis(2016)]{mirjalili2016woa}
Mirjalili, S.; Lewis, A.
The Whale Optimization Algorithm.
{\em Adv. Eng. Softw.} {\bf 2016}, {\em 95},~51--67. [\href{http://dx.doi.org/10.1016/j.advengsoft.2016.01.008}{CrossRef}]
\bibitem[Rana et~al.(2020)Rana, Latiff, Abdulhamid, et~al.]{rana2020whale}
Rana, N.; Latiff, M.S.A.; Abdulhamid, S.I.M.; Chiroma, H.
Whale optimization algorithm: A systematic review of contemporary applications, modifications and developments.
{\em Neural Comput. Appl.} {\bf 2020}, {\em 32},~16245--16277. [\href{http://dx.doi.org/10.1007/s00521-020-04849-z}{CrossRef}]
\bibitem[Salcedo-Sanz et~al.(2014)Salcedo-Sanz, Ser, Landa-Torres, Gil-López, and Portilla-Figueras]{salcedo2014cro}
Salcedo-Sanz, S.; Ser, J.D.; Landa-Torres, I.; Gil-López, S.; Portilla-Figueras, J.
The Coral Reefs Optimization Algorithm: A Novel Metaheuristic for Efficiently Solving Optimization Problems.
{\em Sci. World J.} {\bf 2014}, {\em 2014},~739768. [\href{http://dx.doi.org/10.1155/2014/739768}{CrossRef}]
\bibitem[Faramarzi et~al.(2020)Faramarzi, Heidarinejad, Mirjalili, and Gandomi]{faramarzi2020mpa}
Faramarzi, A.; Heidarinejad, M.; Mirjalili, S.; Gandomi, A.
Marine Predators Algorithm: A nature-inspired metaheuristic.
{\em Expert Syst. Appl.} {\bf 2020}, {\em 152},~113377. [\href{http://dx.doi.org/10.1016/j.eswa.2020.113377}{CrossRef}]
\bibitem[Al-Betar et~al.(2023)Al-Betar, Awadallah, Makhadmeh, et~al.]{albetar2023mpa}
Al-Betar, M.A.; Awadallah, M.A.; Makhadmeh, S.N.; Alyasseri, Z.A.A.; Al-Naymat, G.; Mirjalili, S.
Marine Predators Algorithm: A Review.
{\em Arch. Comput. Methods Eng.} {\bf 2023}, {\em 30},~3405–3435. [\href{http://dx.doi.org/10.1007/s11831-023-09912-1}{CrossRef}]
\bibitem[Chun et~al.(2024)Chun, Hua, Qi, and Yao]{chun2024improved}
Chun, Y.; Hua, X.; Qi, C.; Yao, Y.X.
Improved marine predators algorithm for engineering design optimization problems.
{\em Sci. Rep.} {\bf 2024}, {\em 14},~13000. [\href{http://dx.doi.org/10.1038/s41598-024-63826-x}{CrossRef}]
\bibitem[Liang et~al.(2020)Liang, Fu, Kang, Gao, and Qiang]{liang2020behavior}
Liang, H.; Fu, Y.; Kang, F.; Gao, J.; Qiang, N.
A Behavior-Driven Coordination Control Framework for Target Hunting by UUV Intelligent Swarm.
{\em IEEE Access} {\bf 2020}, {\em 8},~4838--4859. [\href{http://dx.doi.org/10.1109/ACCESS.2019.2962728}{CrossRef}]
\bibitem[Sun and Lv(2025)]{sun2025multi}
Sun, B.; Lv, Z.
Multi-AUV Dynamic Cooperative Path Planning with Hybrid Particle Swarm and Dynamic Window Algorithm in Three-Dimensional Terrain and Ocean Current Environment.
{\em Biomimetics} {\bf 2025}, {\em 10},~536. [\href{http://dx.doi.org/10.3390/biomimetics10080536}{CrossRef}]
\bibitem[Praczyk(2025)]{praczyk2025neural}
Praczyk, T.
Neural leader–follower swarm control algorithm for underwater vehicles.
{\em Neural Comput. Appl.} {\bf 2025}, {\em 37},~19867--19893. [\href{http://dx.doi.org/10.1007/s00521-025-11429-6}{CrossRef}]
\bibitem[Zhang et~al.(2023)Zhang, Wang, Heinrich, Wang, and Dorigo]{zhang2023hybrid}
Zhang, Y.; Wang, S.; Heinrich, M.K.; Wang, X.; Dorigo, M.
3D hybrid formation control of an underwater robot swarm: Switching topologies, unmeasurable velocities, and system constraints.
{\em ISA Trans.} {\bf 2023}, {\em 136},~345--360. [\href{http://dx.doi.org/10.1016/j.isatra.2022.11.014}{CrossRef}] [\href{http://www.ncbi.nlm.nih.gov/pubmed/36509578}{PubMed}]
\bibitem[Ru et~al.(2023)Ru, Hao, Zhang, Xu, and Jia]{ru2023fgpn}
Ru, J.; Hao, D.; Zhang, X.; Xu, H.; Jia, Z.
Research on a hybrid neural network task assignment algorithm for solving multi-constraint heterogeneous autonomous underwater robot swarms.
{\em Front. Neurorobot.} {\bf 2023}, {\em 16},~1055056. [\href{http://dx.doi.org/10.3389/fnbot.2022.1055056}{CrossRef}]
\bibitem[Mu and Gao(2025)]{mu2025coverage}
Mu, X.; Gao, W.
Coverage path planning for multi-AUV considering ocean currents and sonar performance.
{\em Front. Mar. Sci.} {\bf 2025}, {\em 11},~1483122. [\href{http://dx.doi.org/10.3389/fmars.2024.1483122}{CrossRef}]
\bibitem[Zhao et~al.(2022)Zhao, Hu, Feng, Feng, and Su]{zhao2022cooperative}
Zhao, Z.; Hu, Q.; Feng, H.; Feng, X.; Su, W.
A Cooperative Hunting Method for Multi-AUV Swarm in Underwater Weak Information Environment with Obstacles.
{\em J. Mar. Sci. Eng.} {\bf 2022}, {\em 10},~1266. [\href{http://dx.doi.org/10.3390/jmse10091266}{CrossRef}]
\bibitem[Gussen et~al.(2021)Gussen, Laot, Socheleau, Zerr, Le~Mézo, Bourdon, and Le~Berre]{gussen2021optimisation}
Gussen, C.M.G.; Laot, C.; Socheleau, F.X.; Zerr, B.; Le~Mézo, T.; Bourdon, R.; Le~Berre, C.
Optimization of Acoustic Communication Links for a Swarm of AUVs: The COMET and NEMOSENS Examples.
{\em Appl. Sci.} {\bf 2021}, {\em 11},~8200. [\href{http://dx.doi.org/10.3390/app11178200}{CrossRef}]
\bibitem[W\"ubben et~al.(2020)W\"ubben, K\"onsgen, Udugama, Dekorsy, and F\"orster]{wubben2020challenges}
W\"ubben, D.; K\"onsgen, A.; Udugama, A.; Dekorsy, A.; F\"orster, A.
Challenges and Opportunities in Communications for Autonomous Underwater Vehicles. In {\em AI Technology for Underwater Robots}; Kirchner, F., Straube, S., K\"uhn, D., Hoyer, N., Eds.; \hl{Intelligent Systems, Control and Automation: Science and Engineering}; 
Springer: Cham, Switzerland, 2020; Volume~96. [\href{http://dx.doi.org/10.1007/978-3-030-30683-0_7}{CrossRef}]
\bibitem[Quan and Fry(1995)]{quan1995refractive}
Quan, X.; Fry, E.S.
Empirical equation for the index of refraction of seawater.
{\em Appl. Opt.} {\bf 1995}, {\em 34},~3477--3480. [\href{http://dx.doi.org/10.1364/AO.34.003477}{CrossRef}]
\bibitem[Li et~al.(2025)Li, Li, Sun, Fan, and Cui]{li2025recent}
Li, Z.; Li, W.; Sun, K.; Fan, D.; Cui, W.
Recent progress on underwater wireless communication methods and applications.
{\em J. Mar. Sci. Eng.} {\bf 2025}, {\em 13},~1505. [\href{http://dx.doi.org/10.3390/jmse13081505}{CrossRef}]
\bibitem[Vali et~al.(2025)Vali, Michelson, Ghassemlooy, and Noori]{vali2025survey}
Vali, Z.; Michelson, D.; Ghassemlooy, Z.; Noori, H.
A survey of turbulence in underwater optical wireless communications.
{\em Optik} {\bf 2025}, {\em 320},~172126. [\href{http://dx.doi.org/10.1016/j.ijleo.2024.172126}{CrossRef}]
\bibitem[Wang et~al.(2022)Wang, Wang, Liu, Zhu, Wang, Tong, Song, and Zhang]{wang202215}
Wang, T.; Wang, B.; Liu, L.; Zhu, R.; Wang, L.; Tong, C.; Song, Y.; Zhang, P.
15 Mbps underwater wireless optical communications based on acousto-optic modulator and NRZ-OOK modulation.
{\em Opt. Laser Technol.} {\bf 2022}, {\em 150},~107943. [\href{http://dx.doi.org/10.1016/j.optlastec.2022.107943}{CrossRef}]
\bibitem[Busacca et~al.(2024)Busacca, Galluccio, Palazzo, Panebianco, Qi, and Pompili]{busacca2024adaptive}
Busacca, F.; Galluccio, L.; Palazzo, S.; Panebianco, A.; Qi, Z.; Pompili, D.
Adaptive versus predictive techniques in underwater acoustic communication networks.
{\em Comput. Netw.} {\bf 2024}, {\em 252},~110679. [\href{http://dx.doi.org/10.1016/j.comnet.2024.110679}{CrossRef}]
\bibitem[Pal et~al.(2022)Pal, Campagnaro, Ashraf, Rahman, Ashok, and Guo]{pal2022communication}
Pal, A.; Campagnaro, F.; Ashraf, K.; Rahman, M.; Ashok, A.; Guo, H.
Communication for underwater sensor networks: A comprehensive summary.
{\em ACM Trans. Sens. Netw.} {\bf 2022}, {\em 19},~22. [\href{http://dx.doi.org/10.1145/3546827}{CrossRef}]
\bibitem[Lodovisi et~al.(2018)Lodovisi, Loreti, Bracciale, and Betti]{lodovisi2018performance}
Lodovisi, C.; Loreti, P.; Bracciale, L.; Betti, S.
Performance Analysis of Hybrid Optical–Acoustic AUV Swarms for Marine Monitoring.
{\em Future Internet} {\bf 2018}, {\em 10},~65. [\href{http://dx.doi.org/10.3390/fi10070065}{CrossRef}]
\bibitem[Theocharidis and Kavallieratou(2025)]{theocharidis2025underwater}
Theocharidis, T.; Kavallieratou, E.
Underwater communication technologies: A review.
{\em Telecommun. Syst.} {\bf 2025}, {\em 88},~54. [\href{http://dx.doi.org/10.1007/s11235-025-01279-x}{CrossRef}]
\bibitem[Li et~al.(2022)Li, Xu, Zhao, Han, and Yan]{li2022adaptive}
Li, X.; Xu, S.; Zhao, H.; Han, S.; Yan, L.
An adaptive multi-zone geographic routing protocol for underwater acoustic sensor networks.
{\em Wirel. Netw.} {\bf 2022}, {\em 28},~209--223. [\href{http://dx.doi.org/10.1007/s11276-021-02837-2}{CrossRef}]
\bibitem[Liu et~al.(2013)Liu, Qiao, and Ismail]{liu2013covert}
Liu, S.; Qiao, G.; Ismail, A.
Covert underwater acoustic communication using dolphin sounds.
{\em J. Acoust. Soc. Am.} {\bf 2013}, {\em 133}, EL300--EL306. [\href{http://dx.doi.org/10.1121/1.4795219}{CrossRef}]
\bibitem[Khan et~al.(2025)Khan, Rauf, and Jamil]{khan2025hybrid}
Khan, I.; Rauf, M.; Jamil, A.
A Hybrid CT-DEWCA-Based Energy-Efficient Routing Protocol for Data and Storage Nodes in Underwater Acoustic Sensor Networks.
{\em IEEE Access} {\bf 2025}, {\em 13},~91392--91408. [\href{http://dx.doi.org/10.1109/ACCESS.2025.3573089}{CrossRef}]
\bibitem[Osterloh et~al.(2012)Osterloh, Pionteck, and Maehle]{osterloh2012monsun}
Osterloh, C.; Pionteck, T.; Maehle, E.
MONSUN II: A small and inexpensive AUV for underwater swarms.
In Proceedings of the ROBOTIK 2012; 7th German Conference on Robotics, \hl{Munich, Germany, 21--22 May} 2012; pp.~1--6.
\bibitem[Berlinger et~al.(2021)Berlinger, Wulkop, and Nagpal]{berlinger2021bluebot}
Berlinger, F.; Wulkop, P.; Nagpal, R.
Self-Organized Evasive Fountain Maneuvers with a Bioinspired Underwater Robot Collective.
In Proceedings of the 2021 IEEE International Conference on Robotics and Automation (ICRA), \hl{Xi'an, China, 30 May--5 June} May 2021; pp.~9204--9211. [\href{http://dx.doi.org/10.1109/ICRA48506.2021.9561407}{CrossRef}]
\bibitem[Jaffe et~al.(2017)Jaffe, Franks, Roberts, Mirza, Schurgers, Kastner, and Boch]{jaffe2017swarm}
Jaffe, J.S.; Franks, P.J.S.; Roberts, P.L.D.; Mirza, D.; Schurgers, C.; Kastner, R.; Boch, A.
A swarm of autonomous miniature underwater robot drifters for exploring submesoscale ocean dynamics.
{\em Nat. Commun.} {\bf 2017}, {\em 8},~14189. [\href{http://dx.doi.org/10.1038/ncomms14189}{CrossRef}]
\bibitem[Mayberry et~al.(2025)Mayberry, Cai, Yang, Wang, and Zhang]{Mayberry2025}
Mayberry, S.; Cai, J.; Yang, R.; Wang, J.; Zhang, F.
An open-source underwater robotics platform for aquatic research \& exploration.
{\em HardwareX} {\bf 2025}, {\em 24},~e00715. [\href{http://dx.doi.org/10.1016/j.ohx.2025.e00715}{CrossRef}]
\bibitem[Chen et~al.(2025)Chen, Zhou, Hu, and Zhao]{s25206413}
Chen, Y.; Zhou, X.; Hu, W.; Zhao, B.
An Integrated Experimental System for Unmanned Underwater Vehicle Swarm Control.
{\em Sensors} {\bf 2025}, {\em 25},~6413. [\href{http://dx.doi.org/10.3390/s25206413}{CrossRef}]
\bibitem[Zhang et~al.(2023)Zhang, Gao, Tong, Yang, Zhang, Zhang, and Xu]{zhang2023omnidirectional}
Zhang, H.; Gao, Y.; Tong, Z.; Yang, X.; Zhang, Y.; Zhang, C.; Xu, J.
Omnidirectional optical communication system designed for underwater swarm robotics.
{\em Opt. Express} {\bf 2023}, {\em 31},~18630--18644. [\href{http://dx.doi.org/10.1364/OE.490076}{CrossRef}]
\bibitem[Cong et~al.(2021)Cong, Gu, Zhang, and Gao]{cong2021underwater}
Cong, Y.; Gu, C.; Zhang, T.; Gao, Y.
Underwater robot sensing technology: A survey.
{\em Fundam. Res.} {\bf 2021}, {\em 1},~337--345. [\href{http://dx.doi.org/10.1016/j.fmre.2021.03.002}{CrossRef}]
\bibitem[Jiang and Renner(2024)]{jiang2024low}
Jiang, Y.; Renner, B.C.
Low-Cost Underwater Swarm Acoustic Localization: A Review.
{\em IEEE Access} {\bf 2024}, {\em 12},~25779--25796. [\href{http://dx.doi.org/10.1109/ACCESS.2024.3357359}{CrossRef}]
\bibitem[Guo et~al.(2019)Guo, Sun, Wang, and Akyildiz]{guo2019sustainable}
Guo, H.; Sun, Z.; Wang, P.; Akyildiz, I.F.
Sustainable Underwater Robotic Networks using Autonomous Mobile Wireless Energy Transfer.
In Proceedings of the IEEE Global Communications Conference (GLOBECOM), Waikoloa, HI, USA, \hl{9--13 December} 2019; pp.~1--6. [\href{http://dx.doi.org/10.1109/GLOBECOM38437.2019.9013803}{CrossRef}]
\bibitem[Zhang et~al.(2023)Zhang, Ji, Liu, Zhu, and Xu]{zhang2023autonomous}
Zhang, B.; Ji, D.; Liu, S.; Zhu, X.; Xu, W.
Autonomous Underwater Vehicle navigation: A review.
{\em Ocean Eng.} {\bf 2023}, {\em 273},~113861. [\href{http://dx.doi.org/10.1016/j.oceaneng.2023.113861}{CrossRef}]
\bibitem[Cai et~al.(2021)Cai, Xie, Zhang, Lv, and Yang]{cai2021stream}
Cai, W.; Xie, Q.; Zhang, M.; Lv, S.; Yang, J.
Stream-Function Based 3D Obstacle Avoidance Mechanism for Mobile AUVs in the Internet of Underwater Things.
{\em IEEE Access} {\bf 2021}, {\em 9},~142997--143012. [\href{http://dx.doi.org/10.1109/ACCESS.2021.3119594}{CrossRef}]
\bibitem[Wang et~al.(2025)Wang, Liu, and Song]{wang2025actuation}
Wang, P.; Liu, X.; Song, A.
Actuation and Locomotion of Miniature Underwater Robots: A Survey.
{\em Engineering} {\bf 2025}, {\em 51},~195--214. [\href{http://dx.doi.org/10.1016/j.eng.2024.10.022}{CrossRef}]
\bibitem[Ciuccoli et~al.(2024)Ciuccoli, Screpanti, and Scaradozzi]{ciuccoli2024underwater}
Ciuccoli, N.; Screpanti, L.; Scaradozzi, D.
Underwater simulators analysis for digital twinning.
{\em IEEE Access} {\bf 2024}, {\em 12},~34306--34324. [\href{http://dx.doi.org/10.1109/ACCESS.2024.3370443}{CrossRef}]
\bibitem[Szl{\k{e}}g et~al.(2022)Szl{\k{e}}g, Barczyk, Maruszczak, Zieli{\~n}ski, and Szyma{\~n}ska]{szlkeg2022simulation}
Szl{\k{e}}g, P.; Barczyk, P.; Maruszczak, B.; Zieli{\~n}ski, S.; Szyma{\~n}ska, E.
Simulation environment for underwater vehicles testing and training in unity3d.
In Proceedings of the International Conference on Intelligent Autonomous Systems, \hl{Zagreb, Croatia, 13--16 June 2022}; pp.~844--853. [\href{http://dx.doi.org/10.1007/978-3-031-22216-0_56}{CrossRef}]
\bibitem[Iñigo-Blasco et~al.(2012)Iñigo-Blasco, del Rio, Romero-Ternero, Cagigas-Muñiz, and Vicente-Diaz]{inigo2012robotics}
Iñigo-Blasco, P.; del Rio, F.D.; Romero-Ternero, M.C.; Cagigas-Muñiz, D.; Vicente-Diaz, S.
Robotics software frameworks for multi-agent robotic systems development.
{\em Robot. Auton. Syst.} {\bf 2012}, {\em 60},~803--821. [\href{http://dx.doi.org/10.1016/j.robot.2012.02.004}{CrossRef}]
\bibitem[Zhu et~al.(2023)Zhu, Zhang, Liu, Wu, Bai, Ren, and Geng]{zhu2023multiAUVHRL}
Zhu, Z.; Zhang, L.; Liu, L.; Wu, D.; Bai, S.; Ren, R.; Geng, W.
An Efficient Multi-AUV Cooperative Navigation Method Based on Hierarchical Reinforcement Learning.
{\em J. Mar. Sci. Eng.} {\bf 2023}, {\em 11},~1863. [\href{http://dx.doi.org/10.3390/jmse11101863}{CrossRef}]
\bibitem[Khasawneh et~al.(2021)Khasawneh, Altalhi, Kumar, Aggarwal, Kaiwartya, Khalifeh, Al-Khasawneh, and Alarood]{jmse9111219}
Khasawneh, A.M.G.; Altalhi, M.; Kumar, A.; Aggarwal, G.; Kaiwartya, O.; Khalifeh, A.; Al-Khasawneh, M.A.; Alarood, A.A.
An Efficient Void Aware Framework for Enabling Internet of Underwater Things.
{\em J. Mar. Sci. Eng.} {\bf 2021}, {\em 9},~1219. [\href{http://dx.doi.org/10.3390/jmse9111219}{CrossRef}]
\bibitem[Ali et~al.(2024)Ali, Nadeem, Ahmed, Khan, Khan, and Alharbi]{ali2024ilaf}
Ali, S.; Nadeem, M.; Ahmed, S.; Khan, F.; Khan, M.; Alharbi, A.
Acoustic Sensors Data Transmission Integrity and Endurance with {IoT}-Enabled Location-Aware Framework.
{\em PeerJ Comput. Sci.} {\bf 2024}, {\em 10},~e2452. [\href{http://dx.doi.org/10.7717/peerj-cs.2452}{CrossRef}]
\bibitem[Dhurandher et~al.(2012)Dhurandher, Obaidat, and Gupta]{dhurandher2012aquaglomo}
Dhurandher, S.K.; Obaidat, M.S.; Gupta, M.
An Acoustic Communication Based {AQUA-GLOMO} Simulator for Underwater Networks.
{\em Hum.-Centric Comput. Inf. Sci.} {\bf 2012}, {\em 2},~3. [\href{http://dx.doi.org/10.1186/2192-1962-2-3}{CrossRef}]
\bibitem[Zhang et~al.(2024)Zhang, Xu, Wang, Razzaqi, Wang, and Liu]{zhang2024dsua}
Zhang, J.; Xu, B.; Wang, Z.; Razzaqi, A.; Wang, C.; Liu, X.
An Implicit Acoustic Navigation Sensor-Based Method for AUV Cluster Cooperative Navigation and Sensitivity Analysis of DSUA Sensor.
{\em IEEE Sens. J.} {\bf 2024}, {\em 24},~32343--32355. [\href{http://dx.doi.org/10.1109/JSEN.2024.3412764}{CrossRef}]
\bibitem[Chen et~al.(2025)Chen, Hao, Gao, Wang, and Li]{chen2025ftmpc}
Chen, Y.; Hao, S.; Gao, J.; Wang, J.; Li, L.
Fault-Tolerant Model Predictive Control for Autonomous Underwater Vehicles Considering Unknown Disturbances.
{\em J. Mar. Sci. Eng.} {\bf 2025}, {\em 13},~171. [\href{http://dx.doi.org/10.3390/jmse13010171}{CrossRef}]
\bibitem[McCluskey et~al.(2021)McCluskey, Sprogis, London, Bejder, and Loneragan]{mccluskey2021foraging}
McCluskey, S.; Sprogis, K.; London, J.; Bejder, L.; Loneragan, N.
Foraging preferences of an apex marine predator revealed through stomach content and stable isotope analyses.
{\em Glob. Ecol. Conserv.} {\bf 2021}, {\em 25},~e01396. [\href{http://dx.doi.org/10.1016/j.gecco.2020.e01396}{CrossRef}]
\bibitem[Clapham(2000)]{clapham2000humpback}
Clapham, P.
The humpback whale: Seasonal feeding and breeding in a baleen whale. In {\em Cetacean Societies: Field Studies of Dolphins and Whales}; Mann, J., Connor, R., Tyack, P., Whitehead, H., Eds.; University of Chicago Press: Chicago, IL, USA, 2000; pp.~173--196.
\bibitem[Garcia-Hernandez et~al.(2020)Garcia-Hernandez, Salas-Morera, Carmona-Muñoz, Abraham, and Salcedo-Sanz]{garciahernandez2020cro}
Garcia-Hernandez, L.; Salas-Morera, L.; Carmona-Muñoz, C.; Abraham, A.; Salcedo-Sanz, S.
A novel multi-objective Interactive Coral Reefs Optimization algorithm for the Unequal Area Facility Layout Problem.
{\em Swarm Evol. Comput.} {\bf 2020}, {\em 55},~100688.
 [\href{http://dx.doi.org/10.1016/j.swevo.2020.100688}{CrossRef}]
\bibitem[Alyasseri et~al.(2024)Alyasseri, Ali, Al-Betar, Makhadmeh, Jamil, Awadallah, Braik, and Mirjalili]{alyasseri2024woa}
Alyasseri, Z.A.A.; Ali, N.S.; Al-Betar, M.A.; Makhadmeh, S.N.; Jamil, N.; Awadallah, M.A.; Braik, M.; Mirjalili, S.
Recent advances of whale optimization algorithm, its versions and applications. In {\em Advances in Swarm Intelligence: Variations and Adaptations for Optimization Problems}; Elsevier: \hl{Amsterdam, The Netherlands}, 2024; pp.~1--54. [\href{http://dx.doi.org/10.1016/B978-0-32-395365-8.00008-7}{CrossRef}]
\bibitem[Makhadmeh et~al.(2025)Makhadmeh, Kassaymeh, Rjoub, Bataineh, Sanjalawe, and Al-Betar]{makhadmeh2025mowoa}
Makhadmeh, S.N.; Kassaymeh, S.; Rjoub, G.; Bataineh, B.; Sanjalawe, Y.; Al-Betar, M.A.
Recent advances in multi-objective whale optimization algorithm, its versions and applications.
{\em J. King Saud Univ.---Comput. Inf. Sci.} {\bf 2025}, {\em 37},~200. [\href{http://dx.doi.org/10.1007/s44443-025-00184-2}{CrossRef}]
\bibitem[Nadimi-Shahraki et~al.(2023)Nadimi-Shahraki, Zamani, Varzaneh, and Mirjalili]{nadimiShahraki2023woa}
Nadimi-Shahraki, M.H.; Zamani, H.; Varzaneh, Z.A.; Mirjalili, S.
A Systematic Review of the Whale Optimization Algorithm: Theoretical Foundation, Improvements, and Hybridizations.
{\em Arch. Comput. Methods Eng.} {\bf 2023}, {\em 30},~4113--4159. [\href{http://dx.doi.org/10.1007/s11831-023-09928-7}{CrossRef}]
\bibitem[Bujok(2023)]{bujok2023evaluation}
Bujok, P.
Evaluation of Marine Predator Algorithm by Using Engineering Optimisation Problems.
{\em Mathematics} {\bf 2023}, {\em 11},~4716. [\href{http://dx.doi.org/10.3390/math11234716}{CrossRef}]
\bibitem[Pérez-Aracil et~al.(2021)Pérez-Aracil, Camacho-Gómez, Hernández-Díaz, Pereira, Camacho, and Salcedo-Sanz]{perezaracil2021cro}
Pérez-Aracil, J.; Camacho-Gómez, C.; Hernández-Díaz, A.; Pereira, E.; Camacho, D.; Salcedo-Sanz, S.
Memetic coral reefs optimization algorithms for optimal geometrical design of submerged arches.
{\em Swarm Evol. Comput.} {\bf 2021}, {\em 67},~100958. [\href{http://dx.doi.org/10.1016/j.swevo.2021.100958}{CrossRef}]
\bibitem[Guo and Gao(2009)]{guo2009pso}
Guo, S.; Gao, B.
Path-planning optimization of underwater microrobots in 3-D space by PSO Approach.
In Proceedings of the 2009 IEEE International Conference on Robotics and Biomimetics (ROBIO), \hl{Guilin, China,  19--23 December} 2009; pp.~1615--1620. [\href{http://dx.doi.org/10.1109/ROBIO.2009.5420390}{CrossRef}]
\bibitem[Prasath and Kumanan(2022)]{prasath2022abc}
Prasath, R.; Kumanan, T.
Enhanced artificial bee colony approach for the enhancement and classification of underwater images.
{\em Int. J. Comput. Appl.} {\bf 2022}, {\em 44},~433--443. [\href{http://dx.doi.org/10.1080/1206212X.2020.1784591}{CrossRef}]
\bibitem[Wang et~al.(2014)Wang, Guo, Gandomi, Hao, and Wang]{wang2014ckh}
Wang, G.G.; Guo, L.; Gandomi, A.H.; Hao, G.S.; Wang, H.
Chaotic Krill Herd algorithm.
{\em Inf. Sci.} {\bf 2014}, {\em 274},~17--34. [\href{http://dx.doi.org/10.1016/j.ins.2014.02.123}{CrossRef}]
\bibitem[Bastos~Filho et~al.(2008)Bastos~Filho, de Lima~Neto, Lins, Nascimento, and Lima]{bastos2008fss}
Bastos~Filho, C.J.A.; de~Lima~Neto, F.B.; Lins, A.J.C.C.; Nascimento, A.I.S.; Lima, M.P.
A novel search algorithm based on fish school behavior.
In Proceedings of the 2008 IEEE International Conference on Systems, Man and Cybernetics, \hl{Singapore, 12--15 October} 2008; pp.~2646--2651. [\href{http://dx.doi.org/10.1109/ICSMC.2008.4811695}{CrossRef}]
\bibitem[Kaveh and Farhoudi(2013)]{kaveh2013dea}
Kaveh, A.; Farhoudi, N.
A new optimization method: Dolphin echolocation.
{\em Adv. Eng. Softw.} {\bf 2013}, {\em 59},~53--70. [\href{http://dx.doi.org/10.1016/j.advengsoft.2013.03.004}{CrossRef}]
\bibitem[Milner et~al.(2023)Milner, Sooriyabandara, and Hauert]{milner2023swarm}
Milner, E.; Sooriyabandara, M.; Hauert, S.
Swarm Performance Indicators: Metrics for Robustness, Fault Tolerance, Scalability and Adaptability.
{\em arXiv} {\bf 2023}, arXiv:2311.01944. [\href{http://dx.doi.org/10.48550/arXiv.2311.01944}{CrossRef}]
\end{thebibliography}
\end{document}